\documentclass[journal]{IEEEtran}
\ifCLASSINFOpdf
\else
\fi
\usepackage{cite}
\usepackage{amsmath,amssymb,amsfonts}
\usepackage{bm}
\usepackage{algorithmic}
\usepackage{graphicx}
\usepackage{textcomp}
\usepackage{color}

\usepackage{array}

\usepackage{multirow}
\usepackage{threeparttable}
\usepackage{caption2}
\usepackage{booktabs}
\usepackage{graphicx}
\usepackage{floatrow}
\usepackage{soul}

\usepackage[pagebackref=true,breaklinks=true,linkcolor=red,anchorcolor=blue, citecolor=green,letterpaper=true,colorlinks,bookmarks=false]{hyperref}

\makeatletter
  \newcommand\figcaption{\def\@captype{figure}\caption}
  \newcommand\tabcaption{\def\@captype{table}\caption}
\makeatother
\begin{document}
%
\title{Extended Local Binary Patterns for Efficient and Robust Spontaneous Facial Micro-Expression Recognition}
%
%
%


\author{Chengyu~Guo,
Jingyun~Liang,
Geng~Zhan,
Zhong~Liu,
Matti Pietik\"ainen,
and~Li~Liu
\thanks{C. Guo (chloeguo@yeah.net), J. liang (michaelliang12@163.com), Z. Liu (liuzhong@nudt.edu.cn) and L. Liu (li.liu@oulu.fi) was with the College of Systems Engineering, National University of Defense Technology, Changsha 410073, China.}
\thanks{L. Liu and M. Pietik\"ainen (mkp@ee.oulu.fi) was with the Center for Machine Vision and Signal analysis, University of Oulu, Finland.}
\thanks{G. Zhan (geng.zhan@sydney.edu.au) was with School of Electrical and Information Engineering, The University of Sydney, NEW 2006, Australia.}}
\maketitle

\begin{abstract}
Facial Micro-Expressions (MEs) are spontaneous, involuntary facial movements when a person experiences an emotion but deliberately or unconsciously attempts to conceal his or her genuine emotions. Recently, ME recognition has attracted increasing attention due to its potential applications such as clinical diagnosis, business negotiation, interrogations, and security. However, it is expensive to build large scale ME datasets, mainly due to the difficulty of inducing spontaneous MEs. This limits the application of deep learning techniques which require lots of training data.

In this paper, we propose a simple, efficient yet robust descriptor called Extended Local Binary Patterns on Three Orthogonal Planes (ELBPTOP) for ME recognition. ELBPTOP consists of three complementary binary descriptors: LBPTOP and two novel ones Radial Difference LBPTOP (RDLBPTOP) and Angular Difference LBPTOP (ADLBPTOP), which explore the local second order information along the radial and angular directions contained in ME video sequences. ELBPTOP is a novel ME descriptor inspired by unique and subtle facial movements. It is computationally efficient and only marginally increases the cost of computing LBPTOP, yet is extremely effective for ME recognition. In addition, by firstly introducing Whitened Principal Component Analysis (WPCA) to ME recognition, we can further obtain more compact and discriminative feature representations, then achieve significantly computational savings. 

Extensive experimental evaluation on three popular spontaneous ME datasets SMIC, CASME \uppercase\expandafter{\romannumeral2} and SAMM show that our proposed ELBPTOP approach significantly outperforms the previous state-of-the-art on all three single evaluated datasets { and achieves promising results on cross-database recognition.}
Our code will be made available.
\end{abstract}


%
\IEEEpeerreviewmaketitle

 \section{Introduction}
\label{sec:introduction}
Facial Micro-Expressions (MEs) are spontaneous, involuntary facial movements when a person experiences an emotion but deliberately or unconsciously attempts to conceal his or her genuine emotions~\cite{li2013spontaneous,yan2014casme,davison2018samm}. MEs are more likely to occur in high-risk environments because there are more risks to show true emotions~\cite{ekman2003darwin}. Recently, automatic facial ME analysis has attracted increasing attention of affective computing researchers and psychologists because of its potential applications such as clinical diagnosis, business negotiation, interrogations, and security~\cite{wu2010micro,pfister2011recognising}. The study of facial MEs is a well established field in psychology, however, it is a relatively new area from the computer vision perspective with many unsolved and challenging problems~\cite{oh2018survey,martinez2016advances}. There are three main challenges in automatic ME analysis.

(1) \textbf{MEs have a very short duration, local and subtle facial movements.} Compared to ordinary facial expressions, the duration of a ME is usually very short, typically being no more than 500 ms~\cite{yan2013fast}. Besides short duration, MEs also have other unique characteristics such as local and subtle facial movements~\cite{porter2008reading}. Because of these unique characteristics, it is very difficult for human beings to recognize MEs.


(2) \textbf{Lack of large scale spontaneous ME datasets.} Datasets have played a key role in visual recognition problems, especially in the era of deep learning which requires large scale datasets for training~\cite{liu2019bow}. ME analysis is not an exception. However, another challenging issue faced by automatic facial ME analysis is the lack of benchmark datasets (especially large scale ME datasets) due to the difficulties in inducing spontaneous MEs { and labeling them}~\cite{li2013spontaneous,oh2018survey}.  {To the best of our knowledge}\cite{merghani2018review} {, there are eight ME datasets: USF-HD}\cite{shreve2011macro},  {Polikovsky's database}\cite{polikovsky2009facial}, YorkDDT\cite{warren2009detecting},  {SMIC}\cite{li2013spontaneous},  {CASME}\cite{yan2013casme},  {CASME} \uppercase\expandafter{\romannumeral2}\cite{yan2014casme},  {CAS(ME)$^2$}\cite{qu2017cas}, and  {SAMM}\cite{davison2018samm}.  {The first two are posed and not publicly-available. Posed MEs are different from naturally occurring spontaneous MEs significantly. Thus recent works focus on spontaneous ME datasets. } All of the datasets are small. Besides, the emotion categories of the collected samples in these datasets are unevenly distributed, because some emotions are easier to elicit hence they have more samples.

(3) \textbf{Lack of efficient and discriminative feature representations.} Above challenges make ME analysis much harder and more demanding than ordinary facial tasks. Therefore, the extraction of efficient and discriminative feature representations becomes especially important for automatic ME analysis.

In automatic ME analysis, there are mainly two tasks: ME spotting and ME recognition. The former refers to the problem of automatically and accurately locating the temporal interval of a micro-movement in a video sequence,  {where extended versions of SMIC}\cite{li2013spontaneous},  {CAS(ME)$^2$}\cite{qu2017cas}, and  {SAMM}\cite{davison2018samm}  {are widely used}; while the latter is to classify the ME in the video into one of the predefined emotion categories (such as Happiness, Sadness, Surprise, Disgust, etc), where  {SMIC}\cite{li2013spontaneous},  {CASME} \uppercase\expandafter{\romannumeral2}\cite{yan2014casme}, and  {SAMM}\cite{davison2018samm}  {are widely adopted}. ME recognition is the focus of this paper.

Like ordinary facial expression recognition, ME recognition consists of three steps: preprocessing, feature representation and classification~\cite{oh2018survey}. As we discussed previously, the development of powerful feature representations plays a very important role in ME recognition, and thus has been one main focus of research~\cite{ojala2002multiresolution}. Representative feature representation approaches for ME recognition are mainly based on Local Binary Patterns (LBP) ~\cite{ojala1996comparative,ojala2002multiresolution}, Local Phase Quantization (LPQ) ~\cite{ul2012visual}, Histogram of Oriented Gradients (HOG) ~\cite{dalal2005histograms} and Optical Flow (OF)~\cite{horn1981determining}.

Despite these efforts, there is still significant room for improvement towards achieving good performance. The small scale of existing ME datasets and the imbalanced distribution of samples are the primary obstacles to applying existing data hungry deep convolutional neural networks which have brought significant breakthroughs in various visual recognition problems in computer vision due to their ability to learn powerful feature representations directly from raw data. Therefore, state-of-the-art methods for ME recognition are still dominated by traditional handcrafted features like Local Binary Patterns on Three Orthogonal Planes (LBPTOP) ~\cite{zhao2007dynamic}, 3D Gradient Oriented Histogram (HOG 3D)~\cite{polikovsky2009facial} and Histograms of Oriented Optical Flow (HOOF)~\cite{chaudhry2009histograms}.

Due to its prominent advantages such as theoretical simplicity, computational efficiency, and robustness to monotonic grey scale changes, the texture descriptor LBP~\cite{liu2017local} has emerged as one of the most prominent features for face recognition~\cite{ahonen2006face}. Its 3D extension LBPTOP~\cite{zhao2007dynamic} is widely used for facial expression and ME recognition~\cite{oh2018learning}. Many variants of LBP have been proposed to improve robustness, and discriminative power, as summarized in recent surveys~\cite{huang2011local,fernandez2013texture}. However, most LBP variants~\cite{liu2016extended,liu2016median} have not been explored for ME recognition. In other words, in contrast to LBP-based face recognition, LBPTOP type ME recognition is surprisingly underexplored. Moreover, current state-of-the-art ME features like LBPTOP and its variants LBPSIP~\cite{wang2014lbp}, LBPMOP~\cite{wang2015efficient}, STLBP-IP~\cite{huang2015facial}, and STRBP~\cite{huang2017spontaneous} suffer from some drawbacks, such as limited representation power of using only one type of binary feature, limited robustness, and increased computational complexity. 

In this paper, in order to build more discriminative features that can inherit the advantages of LBP type features without suffering the shortcoming of using filters as complemental features~\cite{liu2017local} (i.e., the expensive computation cost), we propose a novel binary feature descriptor named Extended Local Binary Patterns on Three Orthogonal Planes (ELBPTOP) for ME recognition. ELBPTOP is a descriptor that, we argue, nicely balances the three concerns: high distinctiveness, good robustness and low computational cost. In addition, LBPTOP can be considered as a special case of the proposed ELBPTOP descriptor. Our contributions of this paper are summarized as follows.

\begin{itemize}
\item Inspired by the unique texture information of human faces and the subtle intensity variations of local subtle facial movements, the novel ELBPTOP encodes not only the first order information, i.e. the pixel difference information between a central pixel and its neighbours (called Center Pixel Difference Vector, CPDV), but also encodes the second order discriminative information in two directions: the radial direction (Radial Pixel Difference Vector, RPDV) and the angular direction (Angular Pixel Difference Vector, APDV). They are named ADLBPTOP and RDLBPTOP respectively. The proposed ELBPTOP is more effective to capture local, subtle intensity changes and thus delivers stronger discriminative power.

\item To achieve our goal of being computationally efficient while preserving distinctiveness,  we then apply Whitened Principal Component Analysis (WPCA) to get a more compact, robust, and discriminative global descriptor. We are aware of the fact that WPCA has proven to be effective in face recognition. However, we argue that we are the first to apply WPCA to the problem of ME recognition, which has its own unique challenges compared to the extensively studied face recognition problem.

\item We provide extensive experimental evaluation on three popular spontaneous ME datasets CASME \uppercase\expandafter{\romannumeral2}, SMIC, and SAMM to test the effectiveness of the proposed approach, and find that our proposed ELBPTOP approach significantly outperforms previous state-of-the-art on all three evaluated datasets. Our proposed ELBPTOP achieves 73.94\% on CASMEII, which is 6.6\% higher than state-of-the-art on this dataset. More impressively, ELBPTOP increases recognition accuracy from 44.7\% to 63.44\% on the SAMM dataset.
\end{itemize}

Although our method is simple and handcrafted, the very strong quality results obtained on three popular ME datasets in addition with the low computational complexity prove the efficiency of our approach for ME recognition.

The remainder of the paper is organized as follows. Section~\ref{sec:relat} reviews related work in micro-expression recognition and gives a brief outline of LBP and LBPTOP. The main model and more details are represented in Section~\ref{sec:metho}, including the proposed ADLBPTOP and the RDLBPTOP descriptors and our ME recognition scheme. Experimental results are presented in Section~\ref{sec:exper}, leading to conclusions in Section~\ref{sec:concl}.


\section{Related works}
\label{sec:relat}

Feature representation approaches of ME recognition can be divided into two distinct categories: geometric-based and appearance-based~\cite{zeng2008survey} methods. Specifically, geometric-based features describe the face geometry such as the shapes and locations of facial landmarks, so they need precise landmarking and alignment procedures. By contrast, appearance-based features describe the intensity and textural information such as wrinkles and shading changes, and they are more robust to illumination changes and alignment error. Thus, appearance-based feature representation methods, including LBPTOP~\cite{zhao2007dynamic}, HOG 3D~\cite{polikovsky2009facial}, HOOF~\cite{chaudhry2009histograms} and deep learning, have been more popular in ME recognition~\cite{oh2018survey}.

\textbf{LBPTOP variants:} 
Since the pioneering work by Pfister \textit{et al.}~\cite{pfister2011recognising}, LBPTOP has emerged as the most popular approach for spontaneous ME analysis, and quite a few variants have been proposed. 
LBP Six Interception Points (LBPSIP)~\cite{wang2014lbp} is based on three intersecting lines crossing over the center point. 
LBP Mean Orthogonal Planes (LBPMOP)~\cite{wang2015efficient} first computes an average plane for three orthogonal planes, and then computes the LBP on the three orthogonal average planes. By reducing redundant information, LBPSIP and LBPMOP achieved better performance. 
\cite{ben2018learning} explores two effective binary face descriptors: Hot Wheel Patterns~\cite{ben2018learning} and Dual-Cross Patterns~\cite{ding2015multi} and makes use of abundant labelled micro-expressions. 
Besides computing the sign of pixel differences, Spatio-Temporal Completed Local Quantized Patterns (STCLQP)~\cite{huang2016spontaneous} also exploits the complementary components of magnitudes and orientations. Decorrelated Local Spatiotemporal Directional Features (DLSTD)~\cite{wang2014micro} uses Robust Principal Component Analysis (RPCA)~\cite{wright2009robust} to extract subtle emotion information and division of 16 Regions of Interest (ROIs) to utilize the Action Unit (AU) information. Spatio-Temporal Local Radon Binary Pattern (STRBP)~\cite{huang2017spontaneous} uses Radon Transform to obtain robust shape features, while Spatiotemporal Local Binary Pattern with Integral Projection (STLBP-IP)~\cite{huang2015facial} turns to integral projections to preserve shape attributes. 

\textbf{HOOF variants:}  {Histograms of Oriented Optical Flow} (HOOF)~\cite{chaudhry2009histograms} is one of the baseline methods that makes use of optical flow in ME recognition.
Facial Dynamics Map (FDM)~\cite{xu2017microexpression} describes local facial dynamics by extracting principal OF direction of each cuboid. Similarly, ~\cite{liu2016main} designs Main Directional Mean Optical Flow (MDMO) features that utilize the AU information from partitioning facial area into 36 ROIs.
Different from these methods, Consistent Optical Flow Maps~\cite{allaert2017consistent} estimates consistent OF to characterize facial movements, which are calculated from 25 ROIs and the OF of each ROI could be in multiple directions. 
Recently, Bi-Weighted Oriented Optical Flow (BI-WOOF)~\cite{liong2018less} makes use of only the apex frame and the onset frame.  
The majority of OF-based methods need to partition the face area precisely to make use of AU information. This improves the performance but increases the complexity of preprocessing.
\cite{zhao2019improved}  {calculates the LBPTOP and HOOF fusion features for automatic Necessary Morphological Patches (NMPs) extraction which combines the AU-based method and the feature selection method.}

\textbf{HOG 3D variants:} HOG 3D~\cite{polikovsky2009facial} is firstly used to recognize posed MEs and then as a baseline on spontaneous MEs. Its variants, the Histogram of Image  {G}radient Orientation (HIGO)~\cite{li2018towards} ignores the magnitude weighting, hence can suppress the influence of illumination. This makes HIGO become one of the most accurate descriptors at present. 
However, it is worth noting that HOG is an edge-based gradient descriptor. It is sensitive to noise when not being filtered, and the use of low pass filters could lead to the loss of subtle motion change information in ME recognition. Besides, the computation process is time-consuming and cumbersome, resulting in slow speed.

\textbf{Deep learning methods:} \cite{kim2016micro} adopts a shallow network with Convolutional Neural Networks (CNN) and Long Short-Term Memory (LSTM). Other neural networks are explored in Dual Temporal Scale Convolutional Neural Network (DTSCNN)~\cite{peng2017dual}, 3D Flow Convolutional Neural Network (3DFCNN)~\cite{li2018micro}, 3D Spatiotemporal Convolutional Neural Networks (3DCNN)~\cite{reddy2019spontaneous} and Micro-Expression Recognition algorithm using Recurrent CNNs (MER-RCNN)~\cite{xia2018spontaneous}. 
These methods achieve some improvements in ME recognition, but they are still significantly below state-of-the-art handcrafted features, mainly due to lack of large scale ME data.

 {Cross-database ME recognition (CDMER) is a new topic in micro-expression analysis. CDMER considers the large difference of feature distributions existing between the training and testing ME samples in a real scenario to exploit more generalizing approach on different datasets collected by different cameras or under different environments. Besides, a combination of different datasets increased the number of subjects and samples, which is beneficial to the data-driven methods and deep-learning methods. The fundamental works of 1st Micro-Expression Grand Challenge (MEGC2018)}~\cite{yap2018facial} {, 2nd Micro-Expression Grand Challenge (MEGC2019)}~\cite{see2019megc} {, and } \cite{zong2019cross}  {facilitate the development of CDMER. Macro to Micro Transfer Learning}~\cite{peng2018macro}  {utilizes transfer learning to implement CNN from big macro-expression datasets to small ME datasets, ranking top in MEGC2018. Besides transfer learning, }\cite{liu2019neural} {adopt two domain adaptation techniques including adversarial training and expression magnification obtain the best results on the full composite database in MEGC2019. Other methods}~\cite{khor2018enriched,liong2018off,liong2019shallow,zhou2019dual,peng2019novel,van2019capsulenet}  {also show promising performance in cross-database challenges.}

\subsection{LBP and LBPTOP}

LBP was firstly proposed in \cite{ojala1996comparative}, and a completed version was developed in \cite{ojala2002multiresolution}. Later on, it was introduced to face recognition in \cite{ahonen2006face} and its 3D extended version LBPTOP was proposed in \cite{zhao2007dynamic} with application to facial expression analysis. 

\begin{figure}
\begin{center}
\includegraphics[width=1.0\linewidth]{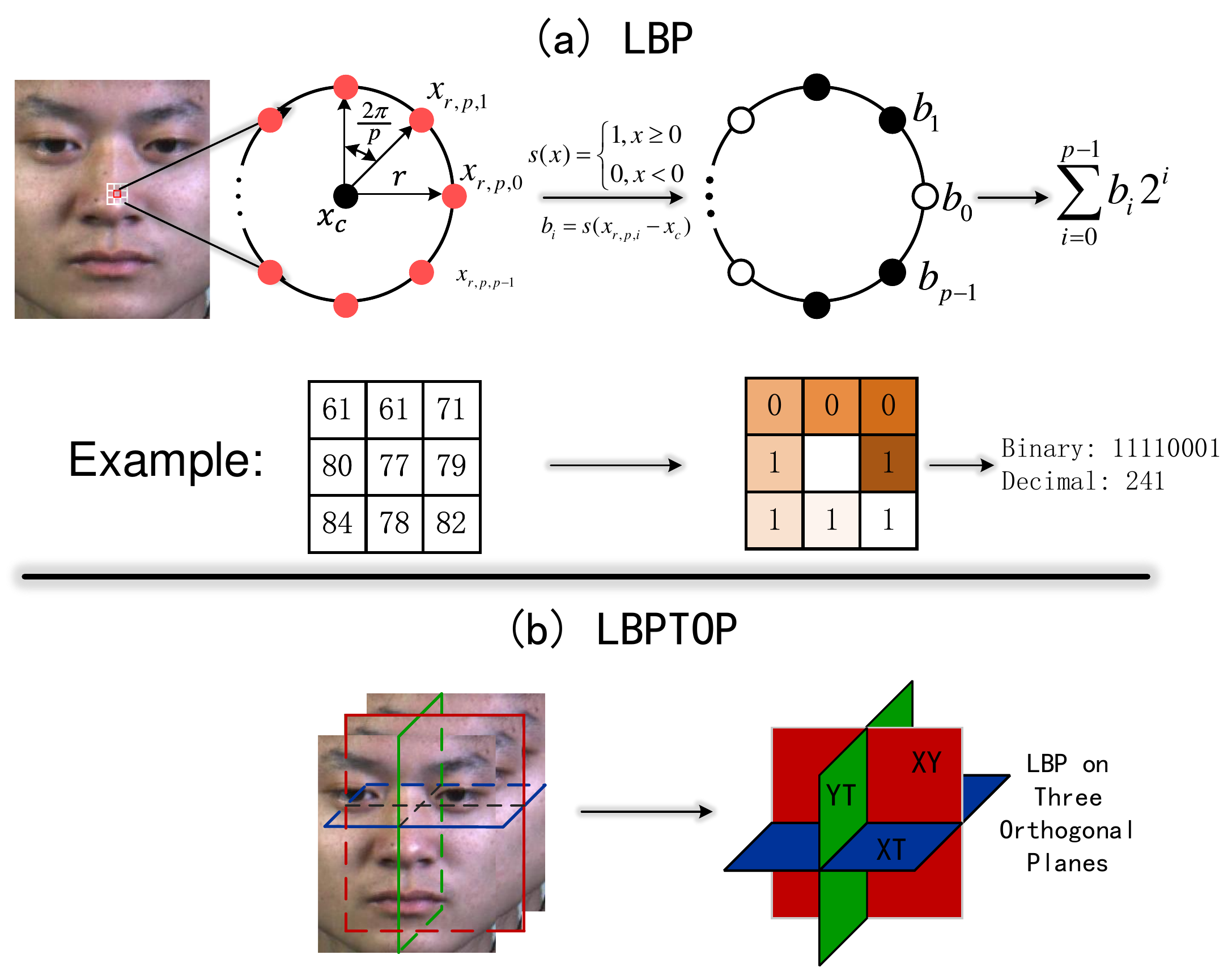}
\end{center}
   \caption{ (a) LBP pattern: The sample neighborhood is the center pixel $x_c$ with $p$ equally spaced pixels on a circle of radius $r$. Then the binary code is calculated by comparing the differences between the center pixel and its neighbors. An example is in the figure. (b) The process of LBPTOP.}
\label{fig:LBP and LBPTOP}
\end{figure}

LBP characterizes the special structure of $p$ pixels, that are evenly distributed in angle on a circle of radius $r$ centered at pixel $x_c$. In specific, as shown in Figure \ref{fig:LBP and LBPTOP}(a), for a central pixel $x_c$  and its $p$ neighboring equally spaced pixels $\{ x_{r,p,n} \}_{n=0}^{p-1}$ on the circle of radius $r$, the LBP pattern is computed via:   

\begin{eqnarray}
\begin{split}
LBP_{r,p}(x_c) = \sum_{n=0}^{p-1}s(x_{r,p,n}-x_c)2^n, s(x)=
\begin{cases}
1,& \text{$x$ $\geq$ 0}\\
0,& \text{$x$ $<$ 0}
\end{cases}
,
\end{split}
\end{eqnarray}
where $s(\cdot)$ is the sign function. The gray values of points that do not fall exactly in the center of pixels are estimated by interpolation. The decimal value of LBP pattern is given by the binary sequence of the circular neighborhood, such as $241=(11110001)_2$ in Figure \ref{fig:LBP and LBPTOP}(a). 
LBP is gray scale invariant and is able to encode important local patterns like lines, edges, and blobs because it measures the differences between the center pixel and its neighbors.


Given an N*M texture image, a LBP pattern $LBP_{r,p}(x_c)$ can be the computed at each pixel c, such that a textured image can be characterized by the distribution of LBP values, representing the whole image by a LBP histogram vector. By altering $r$ and $p$, one can compute LBP features for any quantization of the angular space and for any spatial resolution.

LBPTOP~\cite{zhao2007dynamic} is the 3D extension of LBP by extracting LBP patterns separately from three orthogonal planes: the spatial plane (XY) similar to the regular LBP, the vertical spatiotemporal plane (YT) and the horizontal spatiotemporal plane (XT), as illustrated in Figure \ref{fig:LBP and LBPTOP}(b).

Clearly, LBPTOP encodes temporal changes, and componential information. A video can be represented by concatenating LBP on TOP. Despite a little more complex than the static LBP, LBPTOP can achieve real time processing speed depending on the size of the local sampling neighborhood. The dimensionality of LBPTOP is higher than LBP. Since LBPTOP, which extracts features from TOP, becomes popular when extending 2D spatial appearance descriptors to the spatiotemporal domain.

%
%

\section{Proposed approach}
\label{sec:metho} 
In this section, we first introduce the proposed novel binary descriptor ELBPTOP and then present how to use it for ME recognition.
\subsection{ELBPTOP}
\label{sec:descriptors}

\begin{figure}
\begin{center}
\includegraphics[width=1.0\linewidth]{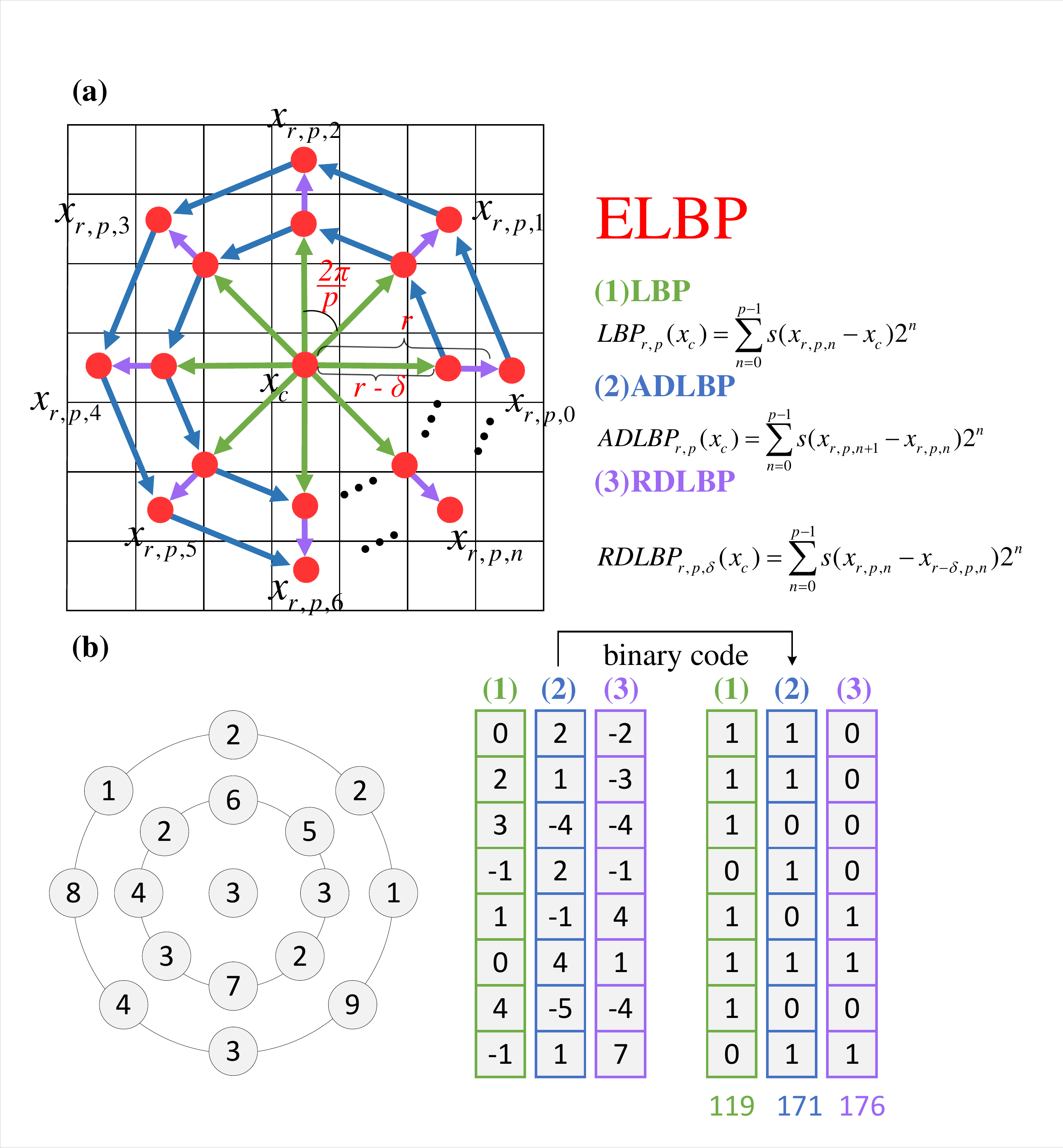}
\end{center}
   \caption{(a) A local circularly symmetric neighbor sampling of ELBP. Two circles of $p=8$ neighbor points are around the central pixel $x_c$. The radius of the inner circle is $r-\delta$, and the radius of the outer circle is $r$. (b) An illustration of the process to calculate ELBP pattern. }
\label{fig:ELBP}
\end{figure}

LBPTOP has emerged as one of the dominant descriptors for ME recognition. Despite this fact, it has several limitations.

\begin{itemize}
\item Currently, LBPTOP~\cite{zhao2007dynamic} usually only exploit the uniform patterns for ME representation. This results in information loss since the proportion of uniform patterns may be too small to capture the variations.
\item It encodes the difference between each pixel and its neighboring pixels only. It is common to combine complementary features like Gabor filters to improve discriminative power. However, this brings a significant computational burden.
\item A large sampling size is helpful since it encodes more local information and provides better representation power. However, increasing the number of sampling points of LBPTOP increases its feature dimensionality significantly.
\end{itemize}

The above analysis leads us to propose novel binary type descriptors, which should not be competitive with LBPTOP, but complement and extend a set of binary feature candidates.

We propose to explore the second order discriminative information in two directions of a local patch: the radial differences (RDLBPTOP) and the angular differences (ADLBPTOP), as complement to the differences between a pixel and its neighbors (LBPTOP). The proposed RDLBPTOP and ADLBPTOP preserve the advantages of LBP, such as a computational efficiency and gray scale invariance.

\begin{figure}
\begin{center}
   \includegraphics[width=1.0\linewidth]{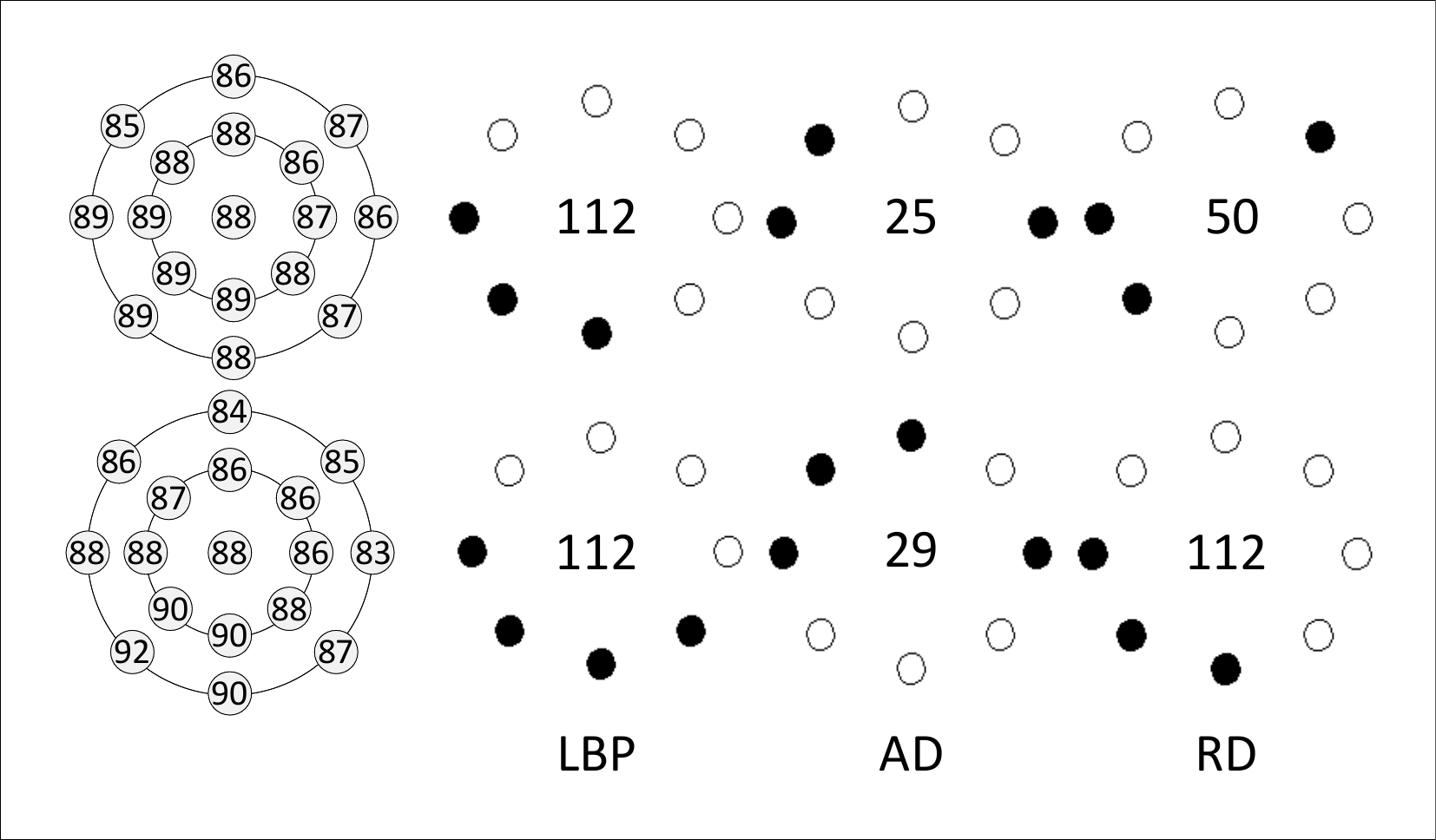}
\end{center}
   \caption{The two given patterns in the left would be considered equivalent by LBP. However, the patterns are, in some ways, quite different from one to others. Fortunately, this underlying change properties can be revealed via angular and radial differences. }
\label{fig:LBP_AD_RD}
\end{figure}

\textbf{(1) Radial Difference Local Binary Pattern (RDLBP)}
As illustrated in Section \ref{sec:relat}, LBP is computed by thresholding the neighboring pixel values on a ring against its center pixel value. It only encodes the relationship between the neighboring pixels on the same ring (i.e. a single scale) and the center one, failing to capture the second order information of neighboring pixels between different rings (different scales). For every pixel in the image, we look at two rings of radii $r$ and $r-\delta$ centered on the pixel $x_c$ and $p$ pixels distributed evenly on each ring, as shown in Figure \ref{fig:ELBP}. To produce the RDLBP codes, we first compute the radial differences $\{ x_{r,p,n}-x_{r-\delta,p,n} \}_n$ between pixels on the two rings and then threshold them against 0. The formal definition of the RDLBP code is as follows:

\begin{eqnarray}
RDLBP_{r,p,\delta}(x_c) = \sum_{n=0}^{p-1}s(x_{r,p,n}-x_{r-\delta,p,n})2^n,
\end{eqnarray}
where $r$ and $r-\delta$ denote the outer ring and the inner ring respectively. As can be seen from Figure~\ref{fig:LBP_AD_RD}, the LBP values of two different pixels can be same in some cases, but for RDLBP,  they are totally different. This is because RDLBP encodes radial pixel difference information.

\begin{figure*}
\begin{center}
   \includegraphics[width=0.7\linewidth]{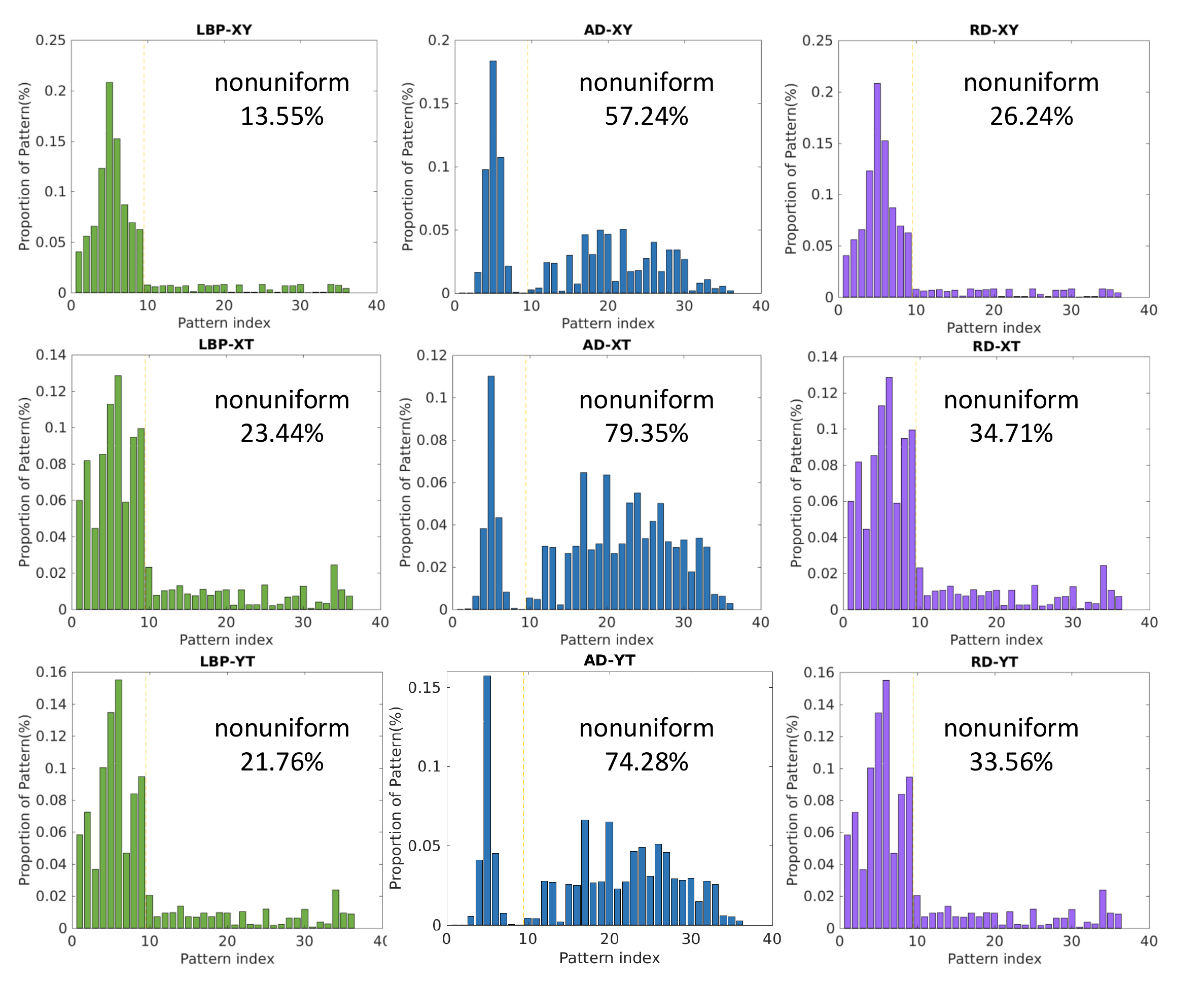}
\end{center}
   \caption{Proportions of the uniform\protect\footnotemark[1] LBPs for the ELBP descriptors (LBP, ADLBP, and RDLBP) on three planes (XY,XT,YT) from the CASME \uppercase\expandafter{\romannumeral2} dataset. The first 9 bins of each histogram are the uniform patterns, and others are the nonuniform patterns. We could observe that the uniform patterns may not account for the major proportion of overall patterns. This is especially obvious in the case of ADLBP. }
\label{fig:scheme}
\end{figure*}

\textbf{(2) Angular Difference Local Binary Pattern (ADLBP)}
LBP also fails to encode the second order information between pixels on the ring. Therefore, ADLBP is composed of neighboring pixel comparisons in angular (like clockwise) direction for all pixels except the center pixel.

Formally, it can be calculated as follows:

\begin{eqnarray}
ADLBP_{r,p}(x_c) = \sum_{n=0}^{p-1}s(x_{r,p,n+1}-x_{r,p,n})2^n.
\end{eqnarray}

Similarly, Figure~\ref{fig:LBP_AD_RD} shows that ADLBP encodes angular difference information, which is different from the original LBP descriptor . It is very compact and provides useful information. We can see that both RDLBP and ADLBP are gray scale invariant and computationally efficient. They can also benefit from rotation invariant extension, uniform extension and 3D extension of LBP.

\textbf{(3) Extended LBP (ELBP)}
We use ELBP to represent the combination of all three binary descriptors: LBP, RDLBP, and ADLBP. The three operators LBP, RDLBP and ADLBP can be combined in two ways, jointly or independently. Because the joint way (3D joint histogram) leads to huge dimension, we use the latter way.

For ME recognition, as shown in Figure \ref{fig:LBP and LBPTOP}(b), we extend ELBP to ELBPTOP. Most LBPTOP based ME descriptors use uniform LBP patterns and group the nonuniform patterns into one bin. However, this leads to lots of information loss because uniform LBPs may not be the majority of LBPs, as illustrated in Figure \ref{fig:scheme}. This is more obvious in the case of ADLBP, where the nonuniform patterns are the dominant patterns. Therefore, in this paper, we use all $2^p$ patterns, rather than uniform patterns only.

\subsection{ELBPTOP for ME recognition}
\label{sec:feature extraction}

\begin{figure}
\begin{center}
   \includegraphics[width=0.8\linewidth]{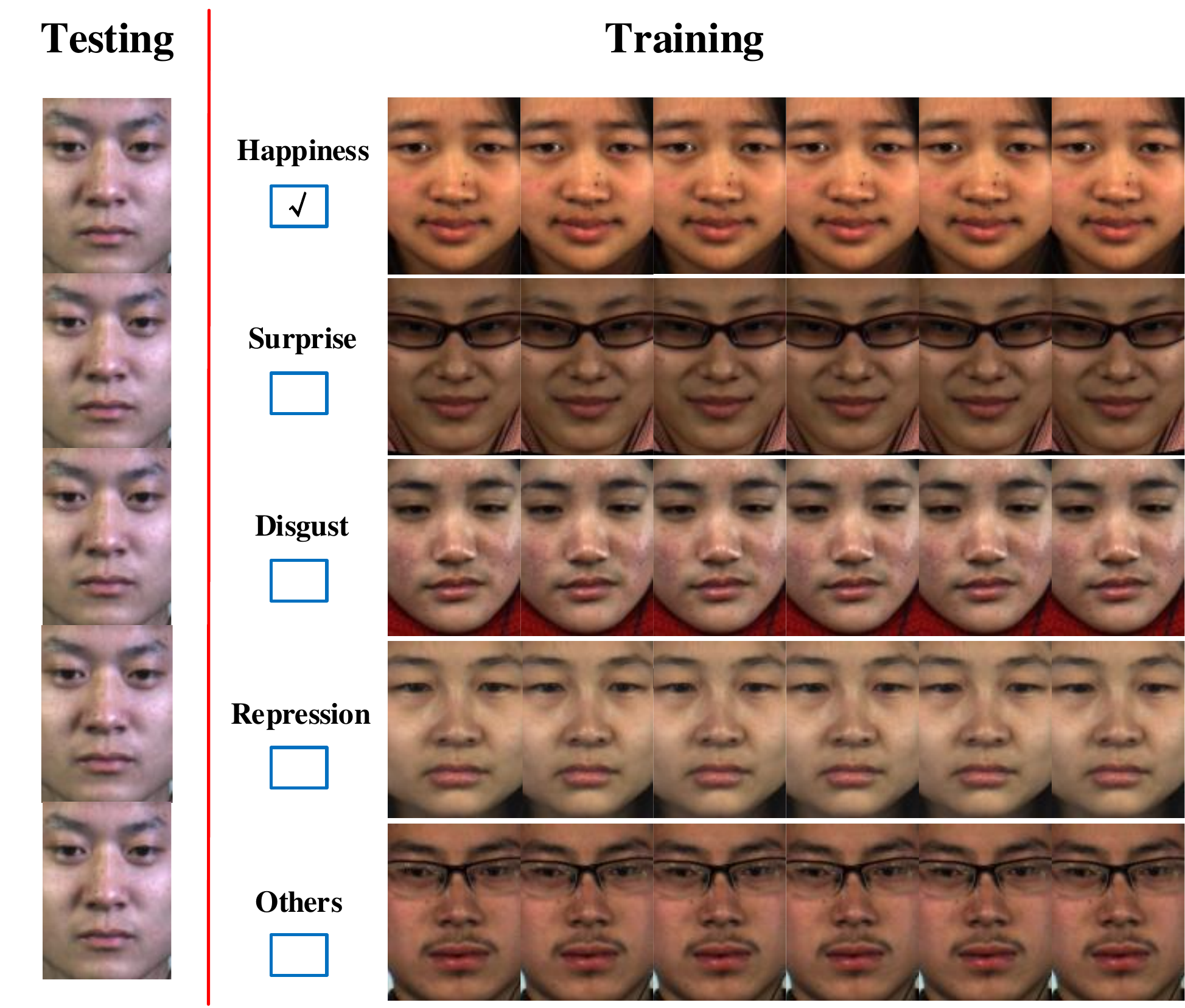}
\end{center}
   \caption{Illustration of the ME classification problem. Samples frames are from CASME \uppercase\expandafter{\romannumeral2}~\cite{yan2014casme}. 
   }
\label{fig:classification}
\end{figure}

\begin{figure*}
\begin{center}
   \includegraphics[width=0.84\linewidth]{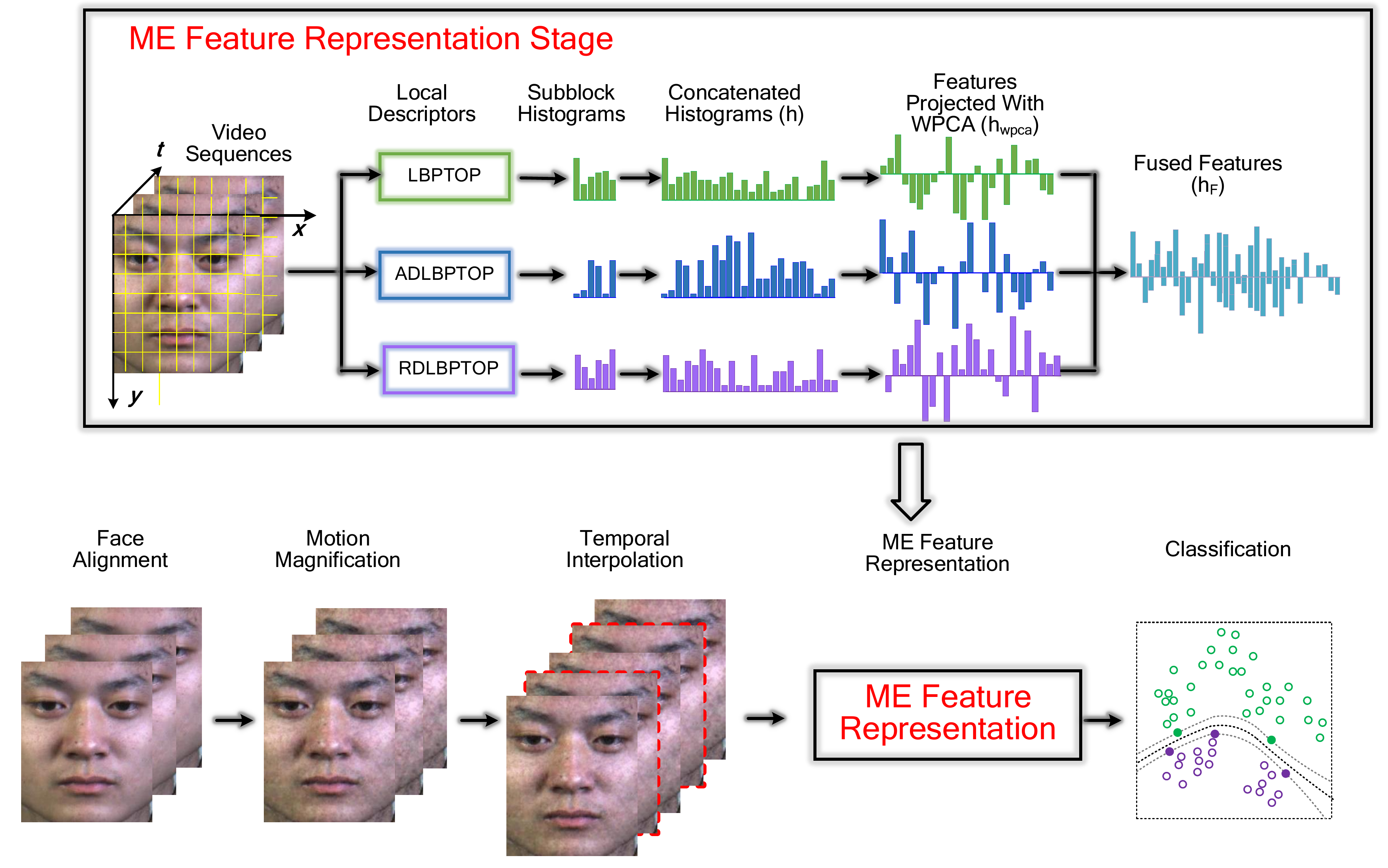}
\end{center}
   \caption{Overview of the proposed ME recognition framework.}
\label{fig:pipeline}
\end{figure*}

In this section, the ME representation is addressed using our proposed ELBPTOP approach to explicitly handle the encountered challenges.

To enhance the discrimination power, we propose to fuse the information extracted by three binary descriptors LBPTOP, RDLBPTOP and ADLBPTOP. The ME feature representation algorithm is illustrated in Figure \ref{fig:pipeline}(a). For each binary descriptor LBPTOP (or RDLBPTOP or ADLBPTOP), the ME video sequences are represented as the concatenated spatiotemporal histograms of the binary codes. In specific, a video sequence is divided into $m \times q \times l$ blocks, then for every single binary descriptor, the dimension of the histogram is $m \times q \times l \times 2^p$. For instance, if we divide the video sequence into $8 \times 8 
\times 2$ blocks, and we choose $p = 8$, the histogram dimension of a single descriptor would be $8 \times 8 
\times 2 \times 2^p = 32768$.

An efficient and effective feature representation scheme is equally important for ME recognition as an efficient and good local descriptor. For each binary code (LBPTOP, RDLBPTOP or ADLBPTOP), the dimension of the feature representation for each ME video sequence is $m \times q \times l \times 2^p$, which is in fact very high. This would cause a computational burden for later classification stage. Therefore, to improve efficiency and preserve distinctiveness, Whitened Principal Component Analysis (WPCA)~\cite{turk1991eigenfaces,nguyen2009local} is firstly introduced for dimensionality reduction before feature fusion.

The idea behind WPCA is that discriminative information is equally distributed along all principle components. The whitening transformation is applied to normalize the contribution of each principal component.
Specifically, given a feature representation $ \boldsymbol h$,
standard PCA is used to get the projected feature $\boldsymbol h_{pca} = \bm {W_{pca}} \boldsymbol h$ at first, where $\bm {W_{pca}}$ is the projection matrix of with $k$ orthonormal columns.
Then, the sorted eigenvectors corresponding to the descending sorted first $k$ principal components are transformed to normalized eigenvectors $\boldsymbol h_{wpca}$ whose variances equal to 1.

In summary, figure \ref{fig:pipeline}(a) illustrates the overview of the proposed feature extraction framework. At first, the video sequences are spatially divided into multiple nonoverlapping subblocks, from each of which three sub-region histograms are extracted via the three proposed binary codes. Each subblock histogram is normalized to sum one. Then, histograms of different subblocks are concatenated and projected by WPCA for dimensionality reduction. Finally, three feature representation vectors with low dimensionality from LBPTOP, RDLBPTOP and ADLBPTOP, are concatenated as a single vector $\boldsymbol h_F$, which is used for final  ME feature representation. 

\subsection{The ME recognition pipeline}

The ME recognition problem is illustrated in Figure \ref{fig:pipeline}(b). The proposed overall pipeline for ME classification is shown in Figure \ref{fig:classification}. Following \cite{li2018towards}, raw ME video sequences are generally processed by the following steps: face alignment, motion magnification, temporal interpolation, feature extraction and classification. 

Our main contribution in this work is the feature representation step, which is presented in detail in the previous sections. A very brief introduction of the other involved steps are given below. Readers are requested to referred to  \cite{li2018towards,pfister2011recognising,davison2018samm} for more information. 
 
\textbf{Face Alignment:} For CASME \uppercase\expandafter{\romannumeral2}~\cite{yan2014casme}, SMIC~\cite{li2013spontaneous} datasets, we use the given cropped images so that face alignment is not required. For SAMM~\cite{davison2018samm} dataset, Active Shape Model~\cite{cootes1995active} is used to detect 77 facial landmarks and then all the facial images are normalized using affine transformation and cropped into the same size according to the eye center points and the outermost points.

\textbf{Motion Magnification:}
Since local intensity changes and facial movement changes in ME are subtle, effective ME characteristics are difficult to capture. To tackle these issues, following \cite{li2018towards,wang2017effective} we use Eulerian Video Magnification (EVM)~\cite{wu2012eulerian} to magnify the subtle motions in videos. 
The goal is to consider the time series of intensity values at any spatial location (pixel) and amplify variation in a given temporal frequency band of interest. The filtered spatial bands are then amplified by a given factor $\alpha$, added back to the original signal, and collapsed to generate the output video. 

\footnotetext[1]{For clear illustration, we transform the ``full'' pattern into ``rotation invariant (ri)'' pattern~\cite{ojala2002multiresolution}. Accordingly, the ``uniform (u2)'' pattern is transformed into ``rotation invariant uniform (riu2)'' pattern. Meanwhile, the proportion of the ``u2" pattern in the "full" pattern is equal to the proportion of the "riu2" pattern in the "ri" pattern. The transformation has no effect on our conclusion. }

\textbf{Temporal Interpolation:}
To address the issue that ME clips are short and have varied duration, we use the Temporal Interpolation Model (TIM) ~\cite{zhou2011towards} and the code provided by \cite{pfister2011recognising}. The model first seeks a low-dimensional manifold where visual features extracted from the frames of a video can be projected onto a continuous deterministic curve embedded in a path graph. Moreover, it can map arbitrary points on the curve back into the image space, making it suitable for temporal interpolation.

\textbf{Classification:}
For classification, we use Linear Support Vector Machine (LSVM)~\cite{chang2011libsvm} as the classifier. Leave-one-subject-out cross-validation (LOSOCV) method is adopted to determine the penalty parameter $c$ in SVM. For each test subject, LOSOCV is applied to the training samples, where in each fold the samples belonging to one subject are served as validation set and the rest of samples compose the new training set to select the best $c$ and the selected $c$ is used for testing.

\section{Experiments}
\label{sec:exper}

\subsection{Datasets}
Three most popular spontaneous datasets, including CASME \uppercase\expandafter{\romannumeral2}~\cite{yan2014casme}, SMIC~\cite{li2013spontaneous} and SAMM~\cite{davison2018samm}, are used to evaluate the performance of the proposed method. The dataset statistics are summarized in Table \ref{dataset}.

\textbf{SMIC~\cite{li2013spontaneous}:} SMIC consists of 164 sample video clips of 16 subjects belonging to 3 different classes , \textit{e.g.}, Positive (51 samples), Negative (70 samples) and Surprise (43 samples). The SMIC data has three versions: a high-speed camera (HS) version at 100 fps, a normal visual camera (VIS) version at 25 fps and a near-infrared camera (NIR) version at 25 fps. The HS camera was used to record all data, while VIS and NIR cameras were only used for the recording of the last eight subjects' data.  {The emotion classes are only based on participants' self-reports.} In this paper, we use the HS samples for experiments, and the resolution of average face size is 160 $\times$ 130. 

\textbf{CASME \uppercase\expandafter{\romannumeral2}~\cite{yan2014casme}:} CASME \uppercase\expandafter{\romannumeral2} contains 247 ME video clips from 26 subjects. All samples are recorded by a high speed camera at 200 fps.  The resolution of samples is 640 $\times$ 480 pixels and the cropped area has 340 $\times$ 280 pixels.  These samples are categorized into five ME classes: Happiness (32 samples), Surprise (25 samples), Disgust (64 samples), Repression (27 samples) and Others (99 samples).  {Different from SMIC, CASME II has AU labels following Facial Action Coding System (FACS).} These classes are used in the whole parameter evaluation and they are used for comparison in Table \ref{tab:compare}. To remove the bias of human reporting, \cite{davison2018objective} reorganized the classes based on AU instead of original estimated emotion classes. Performance on  {the reorganized objective }classes are also reported in Table \ref{tab:SAMM}.

\textbf{SAMM~\cite{davison2018samm}:} SAMM database contains 159 ME video clips from 29 subjects. All samples are recorded by a high speed camera at 200 fps. The resolution of samples is 2040 $\times$ 1088 pixels and the cropped facial area has about 400 $\times$ 400 pixels. These samples are categorized into seven AU based  {objective} classes. Classes \uppercase\expandafter{\romannumeral1}-\uppercase\expandafter{\romannumeral6} are linked with Happiness (24 samples), Surprise (13 samples), Anger (20 samples), Disgust (8 samples), Sadness (3 samples), and Fear (7 samples). Class \uppercase\expandafter{\romannumeral7} (84 samples) relates to contempt and other AUs that have no emotional link in  {FACS}~\cite{ekman1978facial}. We carry on experiment on SAMM with classes \uppercase\expandafter{\romannumeral1}-\uppercase\expandafter{\romannumeral5} and the results are shown in Table \ref{tab:SAMM}.

\begin{table}
\begin{center}
\begin{threeparttable}
\resizebox*{8.5cm}{!}{
\begin{tabular}{|c|c|c|c|}
\hline
Feature & SMIC-HS~\cite{li2013spontaneous} & CASME \uppercase\expandafter{\romannumeral2}~\cite{yan2014casme} & SAMM~\cite{davison2018samm} \\
\hline\hline
No. of Samples & 164 & 247 & 159 \\
No. of Subjects & 16  & 26 & 29 \\
Resolution & 640 $\times$ 480 & 640 $\times$ 480 & 2040 $\times$ 1088 \\
Facial Area & 160 $\times$ 130 & 340 $\times$ 280 &400 $\times$ 400 \\
FPS & 100 & 200 & 200 \\
FACS Coded & NO & Yes & Yes\\
Classes & 3 & 5 & 7 \\
\hline
\end{tabular}
}
\end{threeparttable}
\end{center}
\caption{A summary of the different features of the SMIC, CASME \uppercase\expandafter{\romannumeral2} and SAMM datasets.}
\label{dataset}
\end{table}

\subsection{Implementation details}

 {We conduct two set of experiments: (1) Single database experiment involving SMIC and CASME} \uppercase\expandafter{\romannumeral2}  {with their original estimated emotion classes, and CASME} \uppercase\expandafter{\romannumeral2}  {and SAMM with the reorganized objective classes} \uppercase\expandafter{\romannumeral1}-\uppercase\expandafter{\romannumeral5}.  {(2) Cross-database experiments involving SMIC, CASME } \uppercase\expandafter{\romannumeral2}  {, and SAMM following the Guidelines of the 1st MEGC}~\cite{yap2018facial}  { and the 2nd MEGC}~\cite{see2019megc}.

 {Most of the methods adopt leave-one-subject-out (LOSO) strategy for evaluation. For each fold, all samples from one subject are used as a testing set and the rest for training. A few works}~\cite{wang2014micro,zhao2019improved}  {use leave-one-sample/video-out (LOVO) protocol, in which one sample is used as a testing set and the rest for training. Some works use their own protocols, such as random sampling of test partition}~\cite{ben2018learning,reddy2019spontaneous} {, five-fold}~\cite{xia2018spontaneous} { and others}~\cite{peng2017dual}.
Leave one subject out (LOSO) strategy is used  {for evaluation in all the experiments. Mean accuracy, F1-score, Weighted F1-score, and Unweighted Average Recall (UAR) are used to measure the performance. Mean accuracy is obtained by averaging accuracies of subjects. F1-score is defined as $F = \frac{1}{c}\sum_{i=1}^{c} \frac{2p_i \times r_i}{p_i+r_i}$, where $p_i$ and $r_i$ are the precision and recall of the $i$th ME class, respectively, and $c$ is the number of classes. Weighted F1-scores are weighted by the number of samples in the corresponding classes before averaging. UAR is the ``balanced'' accuracy (averaging the accuracy of each class without consideration of the number of samples per class).}


\textbf{Parameters:}
For block division parameters ($m \times q \times l$), $8\times 8\times 2$ is for CASME \uppercase\expandafter{\romannumeral2} and SMIC, and $5\times 5\times 2$ is for SAMM. For EVM~\cite{wang2017effective}, we choose the second-order bandpass filter with cutoff frequencies $\omega _1=0.4$, $\omega _2=0.05$ and spatial frequency cutoff $\lambda _c=16$. Magnification value $\alpha$ is set to $20$ for CASME \uppercase\expandafter{\romannumeral2} and SAMM, while $\alpha=8$ is chosen for SMIC. TIM~\cite{zhou2011towards} is used to interpolate all ME sequences into the same length 10 according to \cite{li2018towards}. Values of the number of neighboring pixels $p$, outer ring radius $r$ and inner ring radius $r-\delta$ can be found in tables. The WPCA dimension is $v-1$, where $v$ is the number of video clips of each dataset, \textit{e.g.}, 163 for SMIC and 246 for CASME \uppercase\expandafter{\romannumeral2}.

\subsection{Parameter evaluation}
\label{sec:para}

\textbf{The effect of encoding scheme:} 
Table \ref{tab:u2_full} compares the performance of two encoding schemes, full patterns (all $2^P$ patterns) and uniform patterns, on SMIC. Results on single binary descriptor without WPCA are reported. From table \ref{tab:u2_full}, we can see those histogram representations generated by the full patterns significantly outperform the uniform patterns, on all binary descriptors by a large margin (2.22\% to 6.46\%  {in accuracy, and  0.03 to 0.04 in F1-score}), clearly demonstrating the insufficiency of the uniform patterns for representing ME videos. As a result, we conduct rest experiments using the full patterns encoding scheme.

\begin{table}
\begin{center}
\resizebox*{8cm}{!}{
\begin{tabular}{|c|c|c|c|c|}
\hline
\multirow{2}{*}{Method} & \multicolumn{2}{c|}{full patterns} & \multicolumn{2}{c|}{uniform patterns} \\
\cline{2-5}
& Acc. (\%)  {/ F1-score} & $(r,p,\delta)$ & Acc. (\%)  {/ F1-score} & $(r,p,\delta)$  \\
\hline\hline
LBPTOP   & \textbf{52.07} { / \textbf{0.48}}  & (3,8) & 49.85 { / 0.45} & (3,8) \\
ADLBPTOP & \textbf{53.11} { / \textbf{0.51}} & (3,8) & 49.89 { / 0.47} & (3,8)\\
RDLBPTOP & \textbf{53.26} { / \textbf{0.50}} & (3,8,2) & 46.80 { / 0.46} & (3,8,2)\\
\hline
\end{tabular}
}
\end{center}
\caption{ME recognition accuracy (\%) of single descriptors on SMIC using two different encoding schemes: full patterns and uniform patterns. $p$, $r$ and $r-\delta$ indicates the number of neighboring points, the outer ring and the inner ring respectively. All experiments are conducted without WPCA and EVM.}
\label{tab:u2_full}
\end{table}

\begin{table}
\begin{center}
\resizebox*{8.5cm}{!}{
\begin{tabular}{|c|c|c|c|c|c|c|}
\hline
\multirow{2}{*}{Method} & \multicolumn{3}{c|}{original (h)} & \multicolumn{3}{c|}{WPCA (h$^{pca}$)} \\
\cline{2-7}
& Acc. (\%) { /  F1-score} & $(r,p,\delta)$ & Dim. & Acc. (\%) { /  F1-score} & $(r,p,\delta)$ &  Dim.  \\
\hline\hline
LBPTOP    & 51.09 { / 0.47} & (2,8) & 98304 & \textbf{52.29}  { / \textbf{0.49}} & (2,8) & 163\\
ADLBPTOP  & 55.11 { / 0.53}  & (2,8) & 98304 & \textbf{58.45}  { / \textbf{0.54}} & (2,8) & 163\\
RDLBPTOP  & 52.61 { / 0.49} & (2,8,1) & 98304 & \textbf{52.61}  { / \textbf{0.49}} & (2,8,1) & 163\\
\hline
\end{tabular}
}
\end{center}
\caption{ME recognition accuracy (\%) of different binary descriptors on SMIC with or without WPCA. $(m \times q \times l)$ is set to $8\times 8\times 2$. Experiments are conducted without EVM.}
\label{tab:wpca}
\end{table}

\textbf{The effect of WPCA:} Table \ref{tab:wpca} illustrates the effect of WPCA dimensionality reduction on SMIC. Clearly, the accuracy  {and F1-score} of all descriptors is consistently improved by WPCA. Besides, due to much lower feature dimensionality (163 compared with 98304), WPCA could lead to great computational saving. Therefore, further experiments are conducted using WPCA. 

\begin{table}
\begin{center}
\resizebox*{8.5cm}{!}{
\begin{tabular}{|c|c|c|c|c|c|c|}
\hline
\multirow{2}{*}{dataset} & \multicolumn{2}{c|}{ADLBPTOP$_{r,p}$} & \multicolumn{2}{c|}{LBPTOP$_{r,p}$} &\multicolumn{2}{c|}{RDLBPTOP$_{r,p,\delta}$} \\
\cline{2-7}
&Acc. { /  F1-score} & ($r,p$) & Acc. { /  F1-score} & ($r,p$) & Acc. { /  F1-score} & $(r,p,\delta)$ \\
\hline\hline
\multirow{7}{*}{SMIC} &\textbf{62.27} { / \textbf{0.58}} & (1,8) & 52.19 { / \textbf{0.53}} & (1,8) & 52.55 { / \textbf{0.53}} & (1,8,0) \\
& 58.45 { / 0.54} & (2,8) & 52.29 { / 0.49} & (2,8) & 52.61 { / 0.49} & (2,8,1)\\
& 53.11 { / 0.51} & (3,8) & 52.07 { / 0.48} & (3,8)  & 50.67 { / 0.46} & (3,8,1)\\
& 53.11 { / 0.51} & (3,8) & 52.07 { / 0.48} & (3,8)  & 53.26 { / 0.50} & (3,8,2) \\
& 47.95 { / 0.49} & (4,8) & \textbf{54.50} { / 0.53} & (4,8) & \textbf{55.97} { / 0.50} & (4,8,1)\\
& 47.95 { / 0.49} & (4,8) & 54.50 { / 0.53} & (4,8)  & 52.89 { / 0.48} & (4,8,2)\\
& 47.95 { / 0.49} & (4,8) & 54.50 { / 0.53} & (4,8)  & 55.45 { / 0.50} & (4,8,3)\\
\hline
\multirow{7}{*}{CASME \uppercase\expandafter{\romannumeral2}}& 48.35 { / 0.34} & (1,8) & 50.15 { / \textbf{0.38}} & (1,8) & 49.14 { / 0.35} & (1,8,0) \\
& \textbf{56.45} { / \textbf{0.39}} & (2,8) & \textbf{52.79} { / 0.37} & (2,8) & 50.89 { / \textbf{0.38}} & (2,8,1)\\
& 44.36 { / 0.35} & (3,8) & 50.92 { / 0.36} & (3,8) & 49.49 { / 0.35} & (3,8,1)\\
& 44.36 { / 0.35} & (3,8) & 50.92 { / 0.36} & (3,8) & \textbf{55.10} { / 0.37} & (3,8,2) \\
& 47.23 { / 0.35} & (4,8) & 49.49 { / 0.29} & (4,8) & 40.64 { / 0.31} & (4,8,1)\\
& 47.23 { / 0.35} & (4,8) & 49.49 { / 0.29} & (4,8) & 43.19 { / 0.28} & (4,8,2) \\
& 47.23 { / 0.35} & (4,8) & 49.49 { / 0.29} & (4,8) & 45.60 { / 0.32} & (4,8,3) \\
\hline
\end{tabular}
}
\end{center}
\caption{ME recognition accuracy (\%) of the single binary descriptors on SMIC and CASME \uppercase\expandafter{\romannumeral2} under various parameter settings. $(m \times q \times l)$ is set to $8\times 8\times 2$. The WPCA dimension $k$ for SMIC is 163, and 246 for CASME \uppercase\expandafter{\romannumeral2}. Experiments are conducted without EVM.}
\label{tab:SMIC_r}
\end{table}

\textbf{Evaluation of single binary descriptor:} 
To explore the characteristics of different binary descriptors, we conduct experiments under various $(r,p,\delta)$ settings. As shown in Table~\ref{tab:SMIC_r}, the radius $r$ has great impacts on the performance of the three descriptors. The best accuracy often exceeds the second best by a large gap. Therefore, the choice of the best radius $r$ is of great importance.  {It's the same for F1-score. In some cases, the highest accuracy and F1-score do not appear on the same parameters. In the following experiments, we choose the parameter setting with the highest accuracy.} Similarly, $\delta$ is very important for the performance of RDLBP. Comparing the best results of ADLBPTOP, LBPTOP and RDLBPTOP, we can find that the proposed ADLBPTOP and RDLBPTOP outperform LBPTOP on both SMIC and CASME \uppercase\expandafter{\romannumeral2}  {in accuracy}, which shows the importance of radial and angular difference information. Especially, ADLBPTOP performs much better than LBPTOP  {in accuracy and F1-score} (3.66\% and 8.27\% higher  {in accuracy, 0.01 and  0.05 higher in F1-score} on two datasets respectively).

\begin{table}
\begin{center}
\resizebox*{8.5cm}{!}{
\begin{tabular}{|c|c|c|c|c|}
\hline
\multirow{2}{*}{Method} &\multicolumn{2}{c|}{SMIC} & \multicolumn{2}{c|}{CASME \uppercase\expandafter{\romannumeral2}} \\
\cline{2-5}
 & Acc. (\%) { /  F1-score} & $(r,p,\delta)$ & Acc. (\%) { /  F1-score} & $(r,p,\delta)$ \\
\hline\hline
ADLBPTOP  & 62.27 { / 0.58} & (1,8) & 56.45 { / 0.39} & (3,8) \\
\hline
\multirow{2}{*}{ADLBPTOP+EVM} & \textbf{63.73} { / \textbf{0.61}} & (1,8) & 69.12 { / 0.64} & (3,8) \\
& 54.61 { / 0.56} & (4,4) & \textbf{70.20} { / \textbf{0.69}} & (2,4) \\
\hline\hline
LBPTOP & 54.50 { / 0.53} & (4,8) & 52.97 { / 0.37} & (2,8) \\
\hline
\multirow{2}{*}{LBPTOP+EVM} & 60.83 { / \textbf{0.60}} & (3,8) & 67.08 { / 0.58} & (4,8) \\
& \textbf{65.16} { / 0.58} & (3,4) & \textbf{71.55} {\textbf{ / 0.65}} & (3,4) \\
\hline\hline
RDLBPTOP & 55.97 { / 0.51} & (4,8,1) & 55.10 { / 0.37} & (3,8,2) \\
\hline
\multirow{2}{*}{RDLBPTOP+EVM} & 61.04 { / \textbf{0.58}} & (4,8,3) & 67.62 { / \textbf{0.66}} & (3,8,2) \\
& \textbf{62.57} { / 0.56} & (4,4,3) & \textbf{69.24} { / 0.64} & (3,4,1) \\
\hline
\end{tabular}
}
\end{center}
\caption{ME recognition accuracy using different numbers of neighbors $p$ as well as with or without EVM. The parameters of $(m \times q \times l)$ and the WPCA dimensions $k$ are the same as Table \ref{tab:SMIC_r}.}
\label{tab:evm}
\end{table}

\begin{figure*}
\begin{center}
\includegraphics[width=0.82\linewidth]{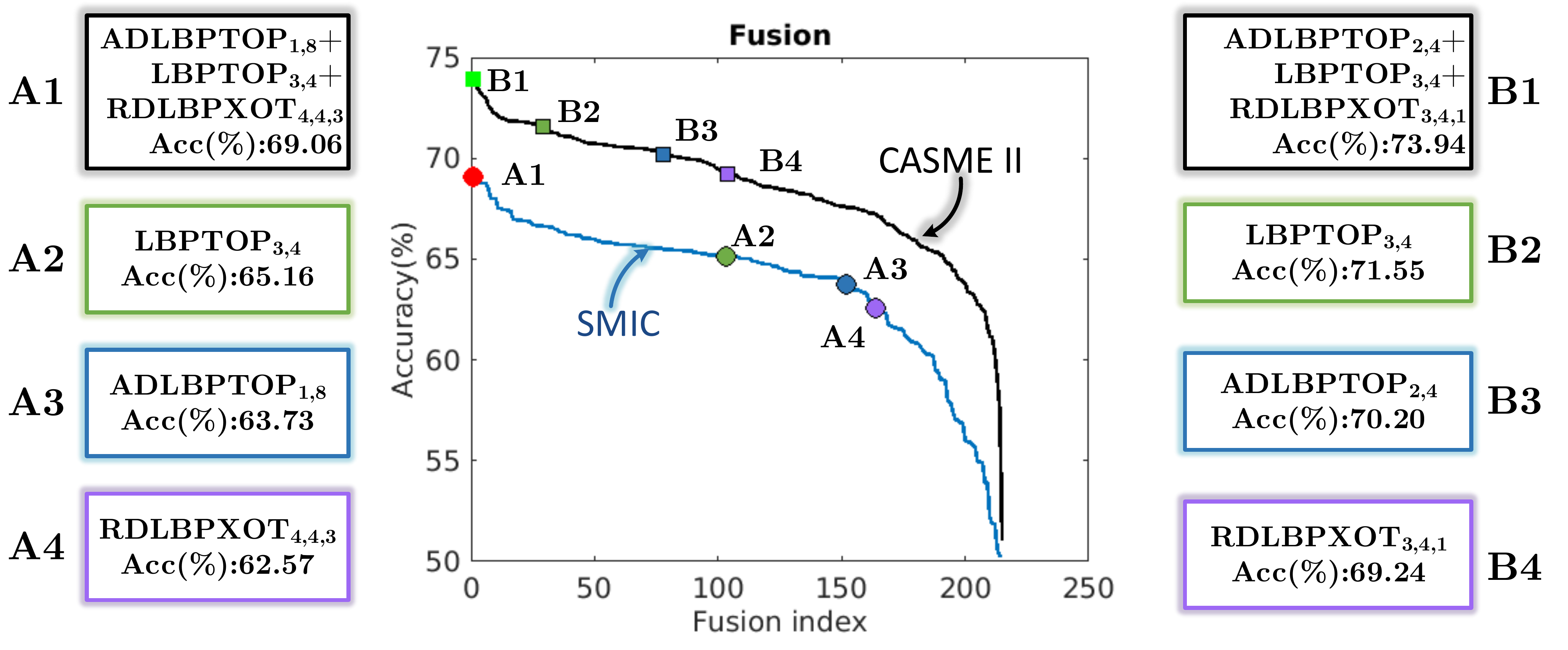}
\end{center}
   \caption{ME recognition accuracy(\%) of different feature fusion schemes on SMIC and CASME \uppercase\expandafter{\romannumeral2}. In the boxes, we show the accuracy of the best fused descriptor and three single binary descriptors.}
\label{fig:fusion}
\end{figure*}

\textbf{Evaluation of EVM and parameter $p$:}
Evaluation of the number of neighboring pixels $p$ and the effect of EVM are summarized in Table \ref{tab:evm}. Note that all the results are reported with their best radii.  We can see that EVM can generally increase the recognition accuracy  {and F1-score}, sometimes significantly (such as for ADLBPTOP and LBPTOP). Table \ref{tab:evm} also indicates that for each single ELBPTOP descriptor, the performance achieved by $p=4$ is better than that by $p=8$, with ADLBPTOP on SMIC being an exception.

\begin{table}
\begin{center}
\resizebox*{8.5cm}{!}{
\begin{tabular}{|c|c|c|c|c|c|}
\hline
\multicolumn{2}{|c|}{\multirow{2}{*}{Method}} &\multicolumn{2}{c|}{SMIC} & \multicolumn{2}{c|}{CASME \uppercase\expandafter{\romannumeral2}} \\
\cline{3-6}
\multicolumn{2}{|c|}{} & Acc. (\%) { / F1-score} & $(r,p,\delta)$ & Acc. (\%) { /  F1-score} & $(r,p,\delta)$ \\
\hline\hline
\multirow{5}{*}{ADLBP} &TOP  & \textbf{63.73} { / \textbf{0.63}} & (1,8) & \textbf{69.12} { / \textbf{0.64}} & (3,8) \\
&XYOT& 55.91 { / 0.57} & (1,8) & 64.12 { / 0.60} & (3,8)  \\
&XOT & 55.92 { / 0.58} & (1,8) & 61.47 { / 0.58} & (3,8) \\
&YOT & 60.22 { / 0.58} & (1,8) & 62.87 { / 0.57} & (3,8) \\
&XY & 55.69 { / 0.51} & (1,8) & 56.46 { / 0.41} & (3,8)  \\
\hline
\multirow{5}{*}{LBP} &TOP & \textbf{60.83} { / 0.59} & (3,8) & \textbf{67.08} { / \textbf{0.61}} & (4,8) \\
&XYOT& 60.47 { / \textbf{0.60}} & (3,8) & 65.04 { / 0.60} & (4,8) \\
&XOT & 57.24 { / 0.58} & (3,8) & 61.38 { / 0.56} & (4,8) \\
&YOT & 55.47 { / 0.57} & (3,8) & 67.65 { / 0.59} & (4,8) \\
&XY & 45.97 { / 0.45} & (3,8) & 60.26 { / 0.43} & (4,8)  \\
\hline
\multirow{5}{*}{RDLBP} &TOP & \textbf{61.04} { / 0.58} & (4,8,3) & 67.62 { / 0.65} & (3,8,2) \\
&XYOT& 57.84 { / \textbf{0.59}} & (4,8,3) & \textbf{68.85} { / \textbf{0.66}} & (3,8,2) \\
&XOT & 56.06 { / 0.56} & (4,8,3) & 62.91 { / 0.60} & (3,8,2) \\
&YOT & 58.76 { / 0.55} & (4,8,3) & 66.56 { / 0.63} & (3,8,2) \\
&XY & 48.85 { / 0.47} & (4,8,3) & 57.14 { / 0.43} & (3,8,2)\\
\hline
\end{tabular}
}
\end{center}
\caption{ME recognition accuracy (\%) of three binary descriptors on different combinations of planes. The parameters $(m \times q \times l)$ and $k$ are the same as Table \ref{tab:SMIC_r}. Experiments are conducted with EVM.
}
\label{tab:fivePlanes}
\end{table}

\textbf{Evaluation of orthogonal planes:} Table \ref{tab:fivePlanes} illustrates the performance of three binary features (LBP, ADLBP, RDLBP) on five combinations of planes.  {TOP, XYOT, XOT, YOT and XY are abbreviations for XY+XT+YT, XT+YT, XT, YT and original spatial plane XY respectively.} It can be observed that TOP  {and XYOT} generally yields the best performance, which indicates that the dynamic information along the time dimension represents the most important information for ME recognition. In contrast, the results on XY plane are almost the worst. This is possibly because the XY plane contains much redundant information about the facial appearance. Maybe not all areas in the facial area contain useful discriminative information for ME recognition.

\textbf{Feature Fusion:} 
In order to find a good fusion of LBP*, RDLBP*, and ADLBP* (here, * represents one of TOP, XYOT, XOT, YOT and XY), we test all 215 ($6^3-1$) possible feature fusion schemes on SMIC and CASME \uppercase\expandafter{\romannumeral2}. All results are shown in Figure \ref{fig:fusion} in descending order. We can see that the highest accuracy is achieved by combining the three type of binary codes. The best results on SMIC-HS is 69.06\%, given by $ADLBPTOP_{1,8}+LBPTOP_{3,4}+RDLBPXOT_{4,4,3}$, and on CASME \uppercase\expandafter{\romannumeral2} is 73.94\%, given by $ADLBPTOP_{2,4}+LBPTOP_{3,4}+RDLBPXOT_{3,4,1}$. 

As it can be seen from Figure \ref{fig:fusion}, that the fused feature increases the accuracy by 3.90\%, 5.33\% and 6.49\% respectively compared with using LBPTOP, ADLBPTOP or RDLBPTOP alone on SMIC. Similarly, the accuracy is improved by 2.39\%,3.74\% and 4.70\% on the three binary codes respectively on CASME \uppercase\expandafter{\romannumeral2}. The strong performance improvement shows that the fused approach indeed captures complementary information. 

\subsection{Comparative evaluation}

\begin{table}
\begin{center}
\begin{threeparttable}
\resizebox*{8.5cm}{!}{
\begin{tabular}{|c|c|c|c|c|}
\hline
\multirow{2}{*}{Method} & \multicolumn{2}{c|}{ {CASME} \uppercase\expandafter{\romannumeral2}} & \multicolumn{2}{c|}{ {SMIC}} \\
\cline{2-5}
&  {Acc. (\%)} &  {F1-score} &  {Acc. (\%)} &  {F1-score} \\
\hline\hline
LBPTOP~\cite{li2013spontaneous}            & --    &-- & 48.78 &-- \\
LBPMOP~\cite{wang2015efficient}            & 44.13 &-- & 50.61 &-- \\
FDM~\cite{xu2017microexpression}           & 45.93\tnote{3} &  {0.41}\tnote{1} & 54.88\tnote{3} &  {0.54}\tnote{1} \\
LBPSIP~\cite{wang2014lbp}                  & 46.56 & -- & 44.51 & -- \\
3DFCNN~\cite{li2018micro}                  & 59.11 &-- & 55.49 &-- \\
STCLQP~\cite{huang2016spontaneous}         & 58.39\tnote{2} &  {0.58} & 64.02\tnote{2} &  {0.64}\\
STLBP-IP~\cite{huang2015facial}            & 59.51 & --& 57.93 & --\\
CNN+LSTM~\cite{kim2016micro}               & 60.98 &-- & --    &--\\
BiWOOF + Phase~\cite{liong2017micro}       & 62.55\tnote{3} & {0.65} & 68.29\tnote{3} & {\textbf{0.67}} \\
Hierahical STLBP-IP~\cite{zong2018learning}& 63.83\tnote{3} & {0.61} & 60.78\tnote{3} & {0.61} \\
STRBP~\cite{huang2017spontaneous}          & 64.37 &-- & 60.98 &-- \\
Discriminative STLBP-IP~\cite{xiaohua2017discriminative} & 64.78 &-- & 63.41 &-- \\
OF Maps~\cite{allaert2017consistent}       & 65.35 &-- & --    &-- \\
HIGOTOP~\cite{li2018towards}               & 67.31 &-- & 68.29 &--\\
ELBPTOP                            & \textbf{73.94} & {\textbf{0.69}} & \textbf{69.06} &  {0.62}\\
\hline
\end{tabular}
}
\begin{tablenotes}
\footnotesize
\item [1]  {The F1-score here is different, which is defined as $F = \frac{2p \times r}{p+r}$, \\
where $p$ and $r$ are the average precision and recall of all the ME class.}
\item [2] Mean recognition rate, which is obtained by averaging accuracies of classes.
\item [3] Overall recognition rate, which is the number of correctly classified \\
samples over the total samples. \\
\end{tablenotes}
\end{threeparttable}
\end{center}
\caption{Comparison between ELBPTOP and previous state-of-the-art methods on CASME \uppercase\expandafter{\romannumeral2} (with original classes) and SMIC. }
\label{tab:compare}
\end{table}

(1) {\textbf{Single database results}}

 {We compare the best results achieved by our ELBPTOP with the baseline method  and  recent and relevant works on CASME }\uppercase\expandafter{\romannumeral2} { and SMIC with their original estimated emotion classes in Table }\ref{tab:compare} {, and  on CASME} \uppercase\expandafter{\romannumeral2}  {and SAMM with the reorganized objective classes in Table }\ref{tab:SAMM}.  {Since the performance with different protocols is quite different, we only compare the methods using the same LOSO strategy. For the same method, results with LOSO are usually lower than those with other protocols (LOVO, k-fold, and so on). }

From Tables \ref{tab:compare} and \ref{tab:SAMM}, we can observe that our proposed approach consistently gives the best results on all three datasets, significantly outperforming the state-of-the-art. As illustrated in Table \ref{tab:compare}, it is clear that our proposed method produces the highest accuracy (73.94\%)  {and the highest F1-score (0.69)}, which is 6.63\% higher  {in accuracy and 0.04 higher in F1-score} than the second best on CASME \uppercase\expandafter{\romannumeral2} (with original classes).
%
%
 {In Table }\ref{tab:SAMM}, our method also surpasses all other methods on CASME \uppercase\expandafter{\romannumeral2} (with reorganized classes) significantly, improving 
  {a margin of 9.91\% in accuracy and 0.10 in F1-score.} The effectiveness of our method is further demonstrated by the large improvement on SAMM, with an increase from 44.70\% to 63.44\% (a margin of 18.74\%). The strong performance on all ME datasets clearly proves that our proposed ELBPTOP is effective for ME recognition.

\begin{table}
\begin{center}
\begin{threeparttable}
\resizebox*{7cm}{!}{
\begin{tabular}{|c|c|c|c|c|}
\hline
\multirow{2}{*}{Method} & \multicolumn{2}{c|}{ {SAMM}} & \multicolumn{2}{c|}{ {CASME} \uppercase\expandafter{\romannumeral2}} \\
\cline{2-5}
&  {Acc. (\%)} &  {F1-score} &  {Acc. (\%)} &  {F1-score} \\
\hline\hline
LBPTOP\cite{davison2018objective}  & 44.70  & {0.35} & 67.80 &  {0.51} \\
 HOOF\cite{davison2018objective}  & 42.17  & {0.33} & 69.64 &  {0.56}  \\
 HOG 3D\cite{davison2018objective}  & 34.16  &  {0.22}& 69.53 &  {0.51}   \\
ELBPTOP  &  \textbf{63.44} & {0.48} & \textbf{79.55} &  {\textbf{0.66}} \\
\hline

\end{tabular}
}
\end{threeparttable}
\end{center}
\caption{Comparison between ELBPTOP and previous state-of-the-art methods on SAMM and CASME \uppercase\expandafter{\romannumeral2} (with reorganized classes). 
}
\label{tab:SAMM}
\end{table}

 {(2) \textbf{Cross database results}}

 {To test the generalization of our method, we also conduct cross database experiments introduced in MEGC2018} \footnote{http://www2.docm.mmu.ac.uk/STAFF/m.yap/FG2018Workshop.htm} {and MEGC2019} \footnote{ https://facial-micro-expressiongc.github.io/MEGC2019/}  {Composite Database Evaluation (CDE) are used to test the performance. Following MEGC2018, all samples from CASME} \uppercase\expandafter{\romannumeral2}  {and SAMM with their reorganized objective classes} \uppercase\expandafter{\romannumeral1}-\uppercase\expandafter{\romannumeral5} { are combined into a single composite database. There are total of 47 subjects (26 from CASME} \uppercase\expandafter{\romannumeral2}  {and 29 from SAMM) and 253 samples (185 from CASME} \uppercase\expandafter{\romannumeral2}  {and 68 from SAMM). The results are shown in Table} \ref{tab:MEGC2018}.  {It can be seen from the table that our method achieves the best F1-score and the second weighted F1-score, confirming the generalization of our method.}            

\begin{table}
\begin{center}
\begin{threeparttable}
\resizebox*{6cm}{!}{
\begin{tabular}{|c|c|c|}
\hline
 {Method} &  {F1-score} &  {Weighted F1-score} \\
\hline\hline
 {HOG 3D}\cite{merghani2018facial}  &  {0.27}  &  {0.44}  \\
 {ELRCN}\cite{khor2018enriched}   &  {0.39}  &  {0.52}  \\
 {LBPTOP}\cite{merghani2018facial}  &  {0.40}  &  {0.52}  \\
 {HOOF}\cite{merghani2018facial}   &  {0.40}  &  {0.53}  \\
 {Transfer learning}\cite{peng2018macro} &  {0.64} & \textbf{ {0.73}} \\
 {ELBPTOP} &  \textbf{ {0.64}} &  {0.71} \\
\hline
\end{tabular}
}
\end{threeparttable}
\end{center}
\caption{The results of composite database evaluation according to MEGC 2018. 
}
\label{tab:MEGC2018}
\end{table}

 {Following MEGC2019, all samples from CASME} \uppercase\expandafter{\romannumeral2}  {and SAMM  and SMIC are combined into a single composite database, and the original emotion classes are grouped into three main classes: negative, positive and surprise. There are total of 68 subjects and 442 samples. Results in Table} \ref{tab:MEGC2019}.  {show that our method is extremely powerful on the CASME} \uppercase\expandafter{\romannumeral2} {, achieving the highest
UAR and F1-score. But there is still a need for further exploration on SMIC and SAMM. We infer that the following factors affect the performance on these two datasets: (1) Different pre-processing methods and cropping areas on SAMM. (2) Big differences in age and ethnicity in SAMM. (3) The lower frame rate and  lower resolution on SMIC.  These factors make the optimal parameters on each data set inconsistent, which in turn affects performance.} 

\begin{table*}
\begin{center}
\begin{threeparttable}
\resizebox*{13cm}{!}{
\begin{tabular}{|c|c|c|c|c|c|c|c|c|}
\hline
\multirow{2}{*}{Method} & \multicolumn{2}{c|}{ {Full}} & \multicolumn{2}{c|}{ {SMIC}} &  \multicolumn{2}{c|}{ {CASME} \uppercase\expandafter{\romannumeral2}} & \multicolumn{2}{c|}{ {SAMM}}\\
\cline{2-9}
&  {F1-score} &  {UAR} &  {F1-score} &  {UAR} &  {F1-score} &  {UAR} &  {F1-score} &  {UAR} \\
\hline\hline
 {LBPTOP}\cite{zhao2007dynamic}  &  {0.59} &  {0.58} &  {0.20} &  {0.53} &  {0.70} &  {0.74} &  {0.40} &  {0.41} \\
 {Bi-WOOF}\cite{liong2018less} &  {0.63} &  {0.62} &  {0.57} &  {0.58} &  {0.78} &  {0.80} &  {0.52} &  {0.51} \\
 {CapsuleNet}\cite{van2019capsulenet} &  {0.65} &  {0.65} &  {0.58} &  {0.59} &  {0.71} &  {0.70} &  {0.62} &  {0.60} \\
 {OFF-ApexNet}\cite{liong2018off} &  {0.72} &  {0.71} &  {0.68} &  {0.67} &  {0.88} &  {0.87} &  {0.54} &  {0.54} \\
 {Dual-Inception Network}\cite{zhou2019dual} &  {0.73} &  {0.73} &  {0.66} &  {0.67} &  {0.86} &  {0.86} &  {0.59} &  {0.57} \\
 {STSTNet}\cite{liong2019shallow} &  {0.74} &  {0.76} &  {0.68} &  {0.70} &  {0.84} &  {0.87} &  {0.66} &  {0.68} \\
 {EMR with Adversarial Training}\cite{liu2019neural} &  {\textbf{0.79}} &  {\textbf{0.78}} &  {\textbf{0.75}} &  {\textbf{0.75}} &  {0.83} &  {0.82} &  {\textbf{0.78}} &  {\textbf{0.72}} \\

 {ELBPTOP}  &  {0.71} &  {0.69} &  {0.65} &  {0.66} &  {\textbf{0.89}} &  {\textbf{0.88}} &  {0.49} &  {0.49}  \\
\hline
\end{tabular}
}
\end{threeparttable}
\end{center}
\caption{The results of composite database evaluation according to MEGC 2019.  
}
\label{tab:MEGC2019}
\end{table*}

\section{Conclusion and future work}
\label{sec:concl}

 In this paper, we proposed a simple, efficient and robust descriptor ELBPTOP for ME recognition. ELBPTOP consists of three complementary binary descriptors: LBPTOP and two novel ones RDLBPTOP and ADLBPTOP, which explore the local second order information along radial and angular directions contained in ME video sequences. For dimension reduction, WPCA is used to obtain efficient and discriminative features. Extensive experiments on three benchmark spontaneous ME datasets, SMIC, CASME \uppercase\expandafter{\romannumeral2} and SAMM have shown that our proposed approach surpasses state-of-the-art by a large margin in single database recognition, and also achieve more promising results on cross-database recognition.
 
  {It is worth noting that there are some difficulties for micro-expression analysis: (1) Lack of standard evaluation  protocol. Different evaluation protocols, performance metrics, number of samples, and emotion classes are chosen by different researchers. It raises the barriers of entry to this topic and increases difficulties for a fair comparison. (2) Lack of large scale spontaneous ME datasets. Small sample size and uneven distribution are still the key to restriction the acquisition of effective features and application to real life. Especially, there are single emotion class in some subject, making it more difficult to obtain features that are distinguishable from expressions rather than distinguishing from subjects. }
 
  {Hand-crafted features confirm that effective discriminant characteristics can be learned. And in the current micro-expression field, many of the deep learning methods are based on hand-crafted features. But some hyper parameters need to be artificially selected, which restricts the performance in cross database problem to some extent. In our future work, we plan to design data-driven methods to learn binary codes directly from data for ME recognition. In addition, in many works, the AU information is very useful but we have not used it in this paper. We will design a better area division algorithm to utilize the AU  information.}

\bibliographystyle{IEEEtran}
\bibliography{egbib}

\begin{thebibliography}{10}
\providecommand{\url}[1]{#1}
\csname url@samestyle\endcsname
\providecommand{\newblock}{\relax}
\providecommand{\bibinfo}[2]{#2}
\providecommand{\BIBentrySTDinterwordspacing}{\spaceskip=0pt\relax}
\providecommand{\BIBentryALTinterwordstretchfactor}{4}
\providecommand{\BIBentryALTinterwordspacing}{\spaceskip=\fontdimen2\font plus
\BIBentryALTinterwordstretchfactor\fontdimen3\font minus
  \fontdimen4\font\relax}
\providecommand{\BIBforeignlanguage}[2]{{%
\expandafter\ifx\csname l@#1\endcsname\relax
\typeout{** WARNING: IEEEtran.bst: No hyphenation pattern has been}%
\typeout{** loaded for the language `#1'. Using the pattern for}%
\typeout{** the default language instead.}%
\else
\language=\csname l@#1\endcsname
\fi
#2}}
\providecommand{\BIBdecl}{\relax}
\BIBdecl

\bibitem{li2013spontaneous}
X.~Li, T.~Pfister, X.~Huang, G.~Zhao, and M.~Pietik{\"a}inen, ``A spontaneous
  micro-expression database: Inducement, collection and baseline,'' in
  \emph{2013 10th IEEE International Conference and Workshops on Automatic Face
  and Gesture Recognition (FG)}.\hskip 1em plus 0.5em minus 0.4em\relax IEEE,
  2013, pp. 1--6.

\bibitem{yan2014casme}
W.-J. Yan, X.~Li, S.-J. Wang, G.~Zhao, Y.-J. Liu, Y.-H. Chen, and X.~Fu,
  ``Casme ii: An improved spontaneous micro-expression database and the
  baseline evaluation,'' \emph{PloS one}, vol.~9, no.~1, p. e86041, 2014.

\bibitem{davison2018samm}
A.~K. Davison, C.~Lansley, N.~Costen, K.~Tan, and M.~H. Yap, ``Samm: A
  spontaneous micro-facial movement dataset,'' \emph{IEEE Transactions on
  Affective Computing}, vol.~9, no.~1, pp. 116--129, 2018.

\bibitem{ekman2003darwin}
P.~Ekman, ``Darwin, deception, and facial expression,'' \emph{Annals of the New
  York Academy of Sciences}, vol. 1000, no.~1, pp. 205--221, 2003.

\bibitem{wu2010micro}
Q.~Wu, X.~Shen, and X.~Fu, ``Micro-expression and its applications,''
  \emph{Advances in Psychological Science}, vol.~18, no.~9, pp. 1359--1368,
  2010.

\bibitem{pfister2011recognising}
T.~Pfister, X.~Li, G.~Zhao, and M.~Pietik{\"a}inen, ``Recognising spontaneous
  facial micro-expressions,'' in \emph{2011 international conference on
  computer vision}.\hskip 1em plus 0.5em minus 0.4em\relax IEEE, 2011, pp.
  1449--1456.

\bibitem{oh2018survey}
Y.-H. Oh, J.~See, A.~C. Le~Ngo, R.~C.-W. Phan, and V.~M. Baskaran, ``A survey
  of automatic facial micro-expression analysis: Databases, methods and
  challenges,'' \emph{Frontiers in psychology}, vol.~9, p. 1128, 2018.

\bibitem{martinez2016advances}
B.~Martinez and M.~F. Valstar, ``Advances, challenges, and opportunities in
  automatic facial expression recognition,'' in \emph{Advances in face
  detection and facial image analysis}.\hskip 1em plus 0.5em minus 0.4em\relax
  Springer, 2016, pp. 63--100.

\bibitem{yan2013fast}
W.-J. Yan, Q.~Wu, J.~Liang, Y.-H. Chen, and X.~Fu, ``How fast are the leaked
  facial expressions: The duration of micro-expressions,'' \emph{Journal of
  Nonverbal Behavior}, vol.~37, no.~4, pp. 217--230, 2013.

\bibitem{porter2008reading}
S.~Porter and L.~Ten~Brinke, ``Reading between the lies: Identifying concealed
  and falsified emotions in universal facial expressions,'' \emph{Psychological
  science}, vol.~19, no.~5, pp. 508--514, 2008.

\bibitem{liu2019bow}
L.~Liu, J.~Chen, P.~Fieguth, G.~Zhao, R.~Chellappa, and M.~Pietik{\"a}inen,
  ``From bow to cnn: Two decades of texture representation for texture
  classification,'' \emph{International Journal of Computer Vision}, vol. 127,
  no.~1, pp. 74--109, 2019.

\bibitem{merghani2018review}
W.~Merghani, A.~K. Davison, and M.~H. Yap, ``A review on facial
  micro-expressions analysis: datasets, features and metrics,'' \emph{arXiv
  preprint arXiv:1805.02397}, 2018.

\bibitem{shreve2011macro}
M.~Shreve, S.~Godavarthy, D.~Goldgof, and S.~Sarkar, ``Macro-and
  micro-expression spotting in long videos using spatio-temporal strain,'' in
  \emph{Face and Gesture 2011}.\hskip 1em plus 0.5em minus 0.4em\relax IEEE,
  2011, pp. 51--56.

\bibitem{polikovsky2009facial}
S.~Polikovsky, Y.~Kameda, and Y.~Ohta, ``Facial micro-expressions recognition
  using high speed camera and 3d-gradient descriptor,'' 2009.

\bibitem{warren2009detecting}
G.~Warren, E.~Schertler, and P.~Bull, ``Detecting deception from emotional and
  unemotional cues,'' \emph{Journal of Nonverbal Behavior}, vol.~33, no.~1, pp.
  59--69, 2009.

\bibitem{yan2013casme}
W.-J. Yan, Q.~Wu, Y.-J. Liu, S.-J. Wang, and X.~Fu, ``Casme database: a dataset
  of spontaneous micro-expressions collected from neutralized faces,'' in
  \emph{2013 10th IEEE international conference and workshops on automatic face
  and gesture recognition (FG)}.\hskip 1em plus 0.5em minus 0.4em\relax IEEE,
  2013, pp. 1--7.

\bibitem{qu2017cas}
F.~Qu, S.-J. Wang, W.-J. Yan, H.~Li, S.~Wu, and X.~Fu, ``Cas (me) $^2$: A
  database for spontaneous macro-expression and micro-expression spotting and
  recognition,'' \emph{IEEE Transactions on Affective Computing}, vol.~9,
  no.~4, pp. 424--436, 2017.

\bibitem{ojala2002multiresolution}
T.~Ojala, M.~Pietik{\"a}inen, and T.~M{\"a}enp{\"a}{\"a}, ``Multiresolution
  gray-scale and rotation invariant texture classification with local binary
  patterns,'' \emph{IEEE Transactions on Pattern Analysis \& Machine
  Intelligence}, no.~7, pp. 971--987, 2002.

\bibitem{ojala1996comparative}
T.~Ojala, M.~Pietik{\"a}inen, and D.~Harwood, ``A comparative study of texture
  measures with classification based on featured distributions,'' \emph{Pattern
  recognition}, vol.~29, no.~1, pp. 51--59, 1996.

\bibitem{ul2012visual}
S.~ul~Hussain and B.~Triggs, ``Visual recognition using local quantized
  patterns,'' in \emph{European conference on computer vision}.\hskip 1em plus
  0.5em minus 0.4em\relax Springer, 2012, pp. 716--729.

\bibitem{dalal2005histograms}
N.~Dalal and B.~Triggs, ``Histograms of oriented gradients for human
  detection,'' 2005.

\bibitem{horn1981determining}
B.~K. Horn and B.~G. Schunck, ``Determining optical flow,'' \emph{Artificial
  intelligence}, vol.~17, no. 1-3, pp. 185--203, 1981.

\bibitem{zhao2007dynamic}
G.~Zhao and M.~Pietikainen, ``Dynamic texture recognition using local binary
  patterns with an application to facial expressions,'' \emph{IEEE Transactions
  on Pattern Analysis \& Machine Intelligence}, no.~6, pp. 915--928, 2007.

\bibitem{chaudhry2009histograms}
R.~Chaudhry, A.~Ravichandran, G.~Hager, and R.~Vidal, ``Histograms of oriented
  optical flow and binet-cauchy kernels on nonlinear dynamical systems for the
  recognition of human actions,'' in \emph{2009 IEEE Conference on Computer
  Vision and Pattern Recognition}.\hskip 1em plus 0.5em minus 0.4em\relax IEEE,
  2009, pp. 1932--1939.

\bibitem{liu2017local}
L.~Liu, P.~Fieguth, Y.~Guo, X.~Wang, and M.~Pietik{\"a}inen, ``Local binary
  features for texture classification: Taxonomy and experimental study,''
  \emph{Pattern Recognition}, vol.~62, pp. 135--160, 2017.

\bibitem{ahonen2006face}
T.~Ahonen, A.~Hadid, and M.~Pietikainen, ``Face description with local binary
  patterns: Application to face recognition,'' \emph{IEEE Transactions on
  Pattern Analysis \& Machine Intelligence}, no.~12, pp. 2037--2041, 2006.

\bibitem{oh2018learning}
T.-H. Oh, R.~Jaroensri, C.~Kim, M.~Elgharib, F.~Durand, W.~T. Freeman, and
  W.~Matusik, ``Learning-based video motion magnification,'' in
  \emph{Proceedings of the European Conference on Computer Vision (ECCV)},
  2018, pp. 633--648.

\bibitem{huang2011local}
D.~Huang, C.~Shan, M.~Ardabilian, Y.~Wang, and L.~Chen, ``Local binary patterns
  and its application to facial image analysis: a survey,'' \emph{IEEE
  Transactions on Systems, Man, and Cybernetics, Part C (Applications and
  Reviews)}, vol.~41, no.~6, pp. 765--781, 2011.

\bibitem{fernandez2013texture}
A.~Fern{\'a}ndez, M.~X. {\'A}lvarez, and F.~Bianconi, ``Texture description
  through histograms of equivalent patterns,'' \emph{Journal of mathematical
  imaging and vision}, vol.~45, no.~1, pp. 76--102, 2013.

\bibitem{liu2016extended}
L.~Liu, P.~Fieguth, G.~Zhao, M.~Pietik{\"a}inen, and D.~Hu, ``Extended local
  binary patterns for face recognition,'' \emph{Information Sciences}, vol.
  358, pp. 56--72, 2016.

\bibitem{liu2016median}
L.~Liu, S.~Lao, P.~W. Fieguth, Y.~Guo, X.~Wang, and M.~Pietik{\"a}inen,
  ``Median robust extended local binary pattern for texture classification,''
  \emph{IEEE Transactions on Image Processing}, vol.~25, no.~3, pp. 1368--1381,
  2016.

\bibitem{wang2014lbp}
Y.~Wang, J.~See, R.~C.-W. Phan, and Y.-H. Oh, ``Lbp with six intersection
  points: Reducing redundant information in lbp-top for micro-expression
  recognition,'' in \emph{Asian conference on computer vision}.\hskip 1em plus
  0.5em minus 0.4em\relax Springer, 2014, pp. 525--537.

\bibitem{wang2015efficient}
------, ``Efficient spatio-temporal local binary patterns for spontaneous
  facial micro-expression recognition,'' \emph{PloS one}, vol.~10, no.~5, p.
  e0124674, 2015.

\bibitem{huang2015facial}
X.~Huang, S.-J. Wang, G.~Zhao, and M.~Piteikainen, ``Facial micro-expression
  recognition using spatiotemporal local binary pattern with integral
  projection,'' in \emph{Proceedings of the IEEE international conference on
  computer vision workshops}, 2015, pp. 1--9.

\bibitem{huang2017spontaneous}
X.~Huang and G.~Zhao, ``Spontaneous facial micro-expression analysis using
  spatiotemporal local radon-based binary pattern,'' in \emph{2017
  International Conference on the Frontiers and Advances in Data Science
  (FADS)}.\hskip 1em plus 0.5em minus 0.4em\relax IEEE, 2017, pp. 159--164.

\bibitem{zeng2008survey}
Z.~Zeng, M.~Pantic, G.~I. Roisman, and T.~S. Huang, ``A survey of affect
  recognition methods: Audio, visual, and spontaneous expressions,'' \emph{IEEE
  transactions on pattern analysis and machine intelligence}, vol.~31, no.~1,
  pp. 39--58, 2008.

\bibitem{ben2018learning}
X.~Ben, X.~Jia, R.~Yan, X.~Zhang, and W.~Meng, ``Learning effective binary
  descriptors for micro-expression recognition transferred by
  macro-information,'' \emph{Pattern Recognition Letters}, vol. 107, pp.
  50--58, 2018.

\bibitem{ding2015multi}
C.~Ding, J.~Choi, D.~Tao, and L.~S. Davis, ``Multi-directional multi-level
  dual-cross patterns for robust face recognition,'' \emph{IEEE transactions on
  pattern analysis and machine intelligence}, vol.~38, no.~3, pp. 518--531,
  2015.

\bibitem{huang2016spontaneous}
X.~Huang, G.~Zhao, X.~Hong, W.~Zheng, and M.~Pietik{\"a}inen, ``Spontaneous
  facial micro-expression analysis using spatiotemporal completed local
  quantized patterns,'' \emph{Neurocomputing}, vol. 175, pp. 564--578, 2016.

\bibitem{wang2014micro}
S.-J. Wang, W.-J. Yan, G.~Zhao, X.~Fu, and C.-G. Zhou, ``Micro-expression
  recognition using robust principal component analysis and local
  spatiotemporal directional features,'' in \emph{European Conference on
  Computer Vision}.\hskip 1em plus 0.5em minus 0.4em\relax Springer, 2014, pp.
  325--338.

\bibitem{wright2009robust}
J.~Wright, A.~Ganesh, S.~Rao, Y.~Peng, and Y.~Ma, ``Robust principal component
  analysis: Exact recovery of corrupted low-rank matrices via convex
  optimization,'' in \emph{Advances in neural information processing systems},
  2009, pp. 2080--2088.

\bibitem{xu2017microexpression}
F.~Xu, J.~Zhang, and J.~Z. Wang, ``Microexpression identification and
  categorization using a facial dynamics map,'' \emph{IEEE Transactions on
  Affective Computing}, vol.~8, no.~2, pp. 254--267, 2017.

\bibitem{liu2016main}
Y.-J. Liu, J.-K. Zhang, W.-J. Yan, S.-J. Wang, G.~Zhao, and X.~Fu, ``A main
  directional mean optical flow feature for spontaneous micro-expression
  recognition,'' \emph{IEEE Transactions on Affective Computing}, vol.~7,
  no.~4, pp. 299--310, 2016.

\bibitem{allaert2017consistent}
B.~Allaert, I.~M. Bilasco, and C.~Djeraba, ``Consistent optical flow maps for
  full and micro facial expression recognition,'' 2017.

\bibitem{liong2018less}
S.-T. Liong, J.~See, K.~Wong, and R.~C.-W. Phan, ``Less is more:
  Micro-expression recognition from video using apex frame,'' \emph{Signal
  Processing: Image Communication}, vol.~62, pp. 82--92, 2018.

\bibitem{zhao2019improved}
Y.~Zhao and J.~Xu, ``An improved micro-expression recognition method based on
  necessary morphological patches,'' \emph{Symmetry}, vol.~11, no.~4, p. 497,
  2019.

\bibitem{li2018towards}
X.~Li, X.~Hong, A.~Moilanen, X.~Huang, T.~Pfister, G.~Zhao, and
  M.~Pietik{\"a}inen, ``Towards reading hidden emotions: A comparative study of
  spontaneous micro-expression spotting and recognition methods,'' \emph{IEEE
  Transactions on Affective Computing}, vol.~9, no.~4, pp. 563--577, 2018.

\bibitem{kim2016micro}
D.~H. Kim, W.~J. Baddar, and Y.~M. Ro, ``Micro-expression recognition with
  expression-state constrained spatio-temporal feature representations,'' in
  \emph{Proceedings of the 24th ACM international conference on
  Multimedia}.\hskip 1em plus 0.5em minus 0.4em\relax ACM, 2016, pp. 382--386.

\bibitem{peng2017dual}
M.~Peng, C.~Wang, T.~Chen, G.~Liu, and X.~Fu, ``Dual temporal scale
  convolutional neural network for micro-expression recognition,''
  \emph{Frontiers in psychology}, vol.~8, p. 1745, 2017.

\bibitem{li2018micro}
J.~Li, Y.~Wang, J.~See, and W.~Liu, ``Micro-expression recognition based on 3d
  flow convolutional neural network,'' \emph{Pattern Analysis and
  Applications}, pp. 1--9, 2018.

\bibitem{reddy2019spontaneous}
S.~P.~T. Reddy, S.~T. Karri, S.~R. Dubey, and S.~Mukherjee, ``Spontaneous
  facial micro-expression recognition using 3d spatiotemporal convolutional
  neural networks,'' \emph{arXiv preprint arXiv:1904.01390}, 2019.

\bibitem{xia2018spontaneous}
Z.~Xia, X.~Feng, X.~Hong, and G.~Zhao, ``Spontaneous facial micro-expression
  recognition via deep convolutional network,'' in \emph{2018 Eighth
  International Conference on Image Processing Theory, Tools and Applications
  (IPTA)}.\hskip 1em plus 0.5em minus 0.4em\relax IEEE, 2018, pp. 1--6.

\bibitem{yap2018facial}
M.~H. Yap, J.~See, X.~Hong, and S.-J. Wang, ``Facial micro-expressions grand
  challenge 2018 summary,'' in \emph{2018 13th IEEE International Conference on
  Automatic Face \& Gesture Recognition (FG 2018)}.\hskip 1em plus 0.5em minus
  0.4em\relax IEEE, 2018, pp. 675--678.

\bibitem{see2019megc}
J.~See, M.~H. Yap, J.~Li, X.~Hong, and S.-J. Wang, ``Megc 2019--the second
  facial micro-expressions grand challenge,'' in \emph{2019 14th IEEE
  International Conference on Automatic Face \& Gesture Recognition (FG
  2019)}.\hskip 1em plus 0.5em minus 0.4em\relax IEEE, 2019, pp. 1--5.

\bibitem{zong2019cross}
Y.~Zong, W.~Zheng, X.~Hong, C.~Tang, Z.~Cui, and G.~Zhao, ``Cross-database
  micro-expression recognition: A benchmark,'' in \emph{Proceedings of the 2019
  on International Conference on Multimedia Retrieval}.\hskip 1em plus 0.5em
  minus 0.4em\relax ACM, 2019, pp. 354--363.

\bibitem{peng2018macro}
M.~Peng, Z.~Wu, Z.~Zhang, and T.~Chen, ``From macro to micro expression
  recognition: deep learning on small datasets using transfer learning,'' in
  \emph{2018 13th IEEE International Conference on Automatic Face \& Gesture
  Recognition (FG 2018)}.\hskip 1em plus 0.5em minus 0.4em\relax IEEE, 2018,
  pp. 657--661.

\bibitem{liu2019neural}
Y.~Liu, H.~Du, L.~Zheng, and T.~Gedeon, ``A neural micro-expression
  recognizer,'' in \emph{2019 14th IEEE International Conference on Automatic
  Face Gesture Recognition (FG 2019)}.\hskip 1em plus 0.5em minus 0.4em\relax
  IEEE, 2019, pp. 1--4.

\bibitem{khor2018enriched}
H.-Q. Khor, J.~See, R.~C.~W. Phan, and W.~Lin, ``Enriched long-term recurrent
  convolutional network for facial micro-expression recognition,'' in
  \emph{2018 13th IEEE International Conference on Automatic Face \& Gesture
  Recognition (FG 2018)}.\hskip 1em plus 0.5em minus 0.4em\relax IEEE, 2018,
  pp. 667--674.

\bibitem{liong2018off}
S.-T. Liong, Y.~Gan, W.-C. Yau, Y.-C. Huang, and T.~L. Ken, ``Off-apexnet on
  micro-expression recognition system,'' \emph{arXiv preprint
  arXiv:1805.08699}, 2018.

\bibitem{liong2019shallow}
S.-T. Liong, Y.~Gan, J.~See, and H.-Q. Khor, ``A shallow triple stream
  three-dimensional cnn (ststnet) for micro-expression recognition system,''
  \emph{arXiv preprint arXiv:1902.03634}, 2019.

\bibitem{zhou2019dual}
L.~Zhou, Q.~Mao, and L.~Xue, ``Dual-inception network for cross-database
  micro-expression recognition,'' in \emph{2019 14th IEEE International
  Conference on Automatic Face \& Gesture Recognition (FG 2019)}.\hskip 1em
  plus 0.5em minus 0.4em\relax IEEE, 2019, pp. 1--5.

\bibitem{peng2019novel}
M.~Peng, C.~Wang, T.~Bi, T.~Chen, X.~Zhou \emph{et~al.}, ``A novel apex-time
  network for cross-dataset micro-expression recognition,'' \emph{arXiv
  preprint arXiv:1904.03699}, 2019.

\bibitem{van2019capsulenet}
N.~Van~Quang, J.~Chun, and T.~Tokuyama, ``Capsulenet for micro-expression
  recognition,'' in \emph{2019 14th IEEE International Conference on Automatic
  Face \& Gesture Recognition (FG 2019)}.\hskip 1em plus 0.5em minus
  0.4em\relax IEEE, 2019, pp. 1--7.

\bibitem{turk1991eigenfaces}
M.~Turk and A.~Pentland, ``Eigenfaces for recognition,'' \emph{Journal of
  cognitive neuroscience}, vol.~3, no.~1, pp. 71--86, 1991.

\bibitem{nguyen2009local}
H.~V. Nguyen, L.~Bai, and L.~Shen, ``Local gabor binary pattern whitened pca: A
  novel approach for face recognition from single image per person,'' in
  \emph{International conference on biometrics}.\hskip 1em plus 0.5em minus
  0.4em\relax Springer, 2009, pp. 269--278.

\bibitem{cootes1995active}
T.~F. Cootes, C.~J. Taylor, D.~H. Cooper, and J.~Graham, ``Active shape
  models-their training and application,'' \emph{Computer vision and image
  understanding}, vol.~61, no.~1, pp. 38--59, 1995.

\bibitem{wang2017effective}
Y.~Wang, J.~See, Y.-H. Oh, R.~C.-W. Phan, Y.~Rahulamathavan, H.-C. Ling, S.-W.
  Tan, and X.~Li, ``Effective recognition of facial micro-expressions with
  video motion magnification,'' \emph{Multimedia Tools and Applications},
  vol.~76, no.~20, pp. 21\,665--21\,690, 2017.

\bibitem{wu2012eulerian}
H.-Y. Wu, M.~Rubinstein, E.~Shih, J.~Guttag, F.~Durand, and W.~Freeman,
  ``Eulerian video magnification for revealing subtle changes in the world,''
  2012.

\bibitem{zhou2011towards}
Z.~Zhou, G.~Zhao, and M.~Pietik{\"a}inen, ``Towards a practical lipreading
  system,'' in \emph{CVPR 2011}.\hskip 1em plus 0.5em minus 0.4em\relax IEEE,
  2011, pp. 137--144.

\bibitem{chang2011libsvm}
C.-C. Chang and C.-J. Lin, ``Libsvm: a library for support vector machines,''
  \emph{ACM transactions on intelligent systems and technology (TIST)}, vol.~2,
  no.~3, p.~27, 2011.

\bibitem{davison2018objective}
A.~Davison, W.~Merghani, and M.~Yap, ``Objective classes for micro-facial
  expression recognition,'' \emph{Journal of Imaging}, vol.~4, no.~10, p. 119,
  2018.

\bibitem{ekman1978facial}
P.~Ekman and W.~V. Friesen, \emph{Facial action coding system: Investigator's
  guide}.\hskip 1em plus 0.5em minus 0.4em\relax Consulting Psychologists
  Press, 1978.

\bibitem{liong2017micro}
S.-T. Liong and K.~Wong, ``Micro-expression recognition using apex frame with
  phase information,'' in \emph{2017 Asia-Pacific Signal and Information
  Processing Association Annual Summit and Conference (APSIPA ASC)}.\hskip 1em
  plus 0.5em minus 0.4em\relax IEEE, 2017, pp. 534--537.

\bibitem{zong2018learning}
Y.~Zong, X.~Huang, W.~Zheng, Z.~Cui, and G.~Zhao, ``Learning from hierarchical
  spatiotemporal descriptors for micro-expression recognition,'' \emph{IEEE
  Transactions on Multimedia}, vol.~20, no.~11, pp. 3160--3172, 2018.

\bibitem{xiaohua2017discriminative}
H.~Xiaohua, S.-J. Wang, X.~Liu, G.~Zhao, X.~Feng, and M.~Pietikainen,
  ``Discriminative spatiotemporal local binary pattern with revisited integral
  projection for spontaneous facial micro-expression recognition,'' \emph{IEEE
  Transactions on Affective Computing}, 2017.

\bibitem{merghani2018facial}
W.~Merghani, A.~Davison, and M.~Yap, ``Facial micro-expressions grand challenge
  2018: evaluating spatio-temporal features for classification of objective
  classes,'' in \emph{2018 13th IEEE International Conference on Automatic Face
  \& Gesture Recognition (FG 2018)}.\hskip 1em plus 0.5em minus 0.4em\relax
  IEEE, 2018, pp. 662--666.

\end{thebibliography}

\end{document}